%% file: main.tex
\documentclass[10pt,twocolumn,letterpaper]{article}

\usepackage[pagenumbers]{cvpr}      
\usepackage{comment}
\usepackage{graphicx}
\usepackage{multirow}
\usepackage{stmaryrd}

\newcommand{\Point}{\mathbf{p}}

\input{preamble}

\definecolor{cvprblue}{rgb}{0.21,0.49,0.74}
\usepackage[pagebackref,breaklinks,colorlinks,citecolor=cvprblue]{hyperref}

\title{Q-REG: End-to-End Trainable Point Cloud Registration with Surface Curvature}

\author{
Shengze Jin$^{1}$ \quad Daniel Barath$^{1}$ \quad Marc Pollefeys$^{1,3}$ \quad Iro Armeni$^{1,2}$\\
$^{1}$ETH Zurich \quad
$^{2}$Stanford University \quad
$^{3}$Microsoft\\
}

\begin{document}
\maketitle
\input{sec/0_abstract}    
\input{sec/1_intro}

\input{sec/2_related_work}
\input{sec/3_method}

\input{sec/4_experiments}

\input{sec/5_conclusion}

{
    \small
    \bibliographystyle{ieeenat_fullname}
    \bibliography{main}
}
\input{sec/X_supplementary}

\end{document}

%% file: preamble.tex
\usepackage[dvipsnames]{xcolor}

%% file: sec/0_abstract.tex
\begin{abstract}

Point cloud registration has seen recent success with several learning-based methods that focus on correspondence matching, and as such, optimize only for this objective. Following the learning step of correspondence matching, they evaluate the estimated rigid transformation with a RANSAC-like framework.
While it is an indispensable component of these methods, it prevents a fully end-to-end training, leaving the objective to minimize the pose error non-served. We present a novel solution, Q-REG, which utilizes rich geometric information to estimate the rigid pose from a single correspondence. Q-REG allows to formalize the robust estimation as an exhaustive search, hence enabling end-to-end training that optimizes over both objectives of correspondence matching and rigid pose estimation. We demonstrate in the experiments that Q-REG is agnostic to the correspondence matching method and provides consistent improvement both when used only in inference and in end-to-end training. It sets a new state-of-the-art on the 3DMatch, KITTI and ModelNet benchmarks.

\end{abstract}

%% file: sec/1_intro.tex
\section{Introduction}

\label{sec:intro}

Point cloud registration is the task of estimating the rigid transformation that aligns two partially overlapping point clouds. 
It is commonly solved by establishing a set of tentative correspondences between the two point clouds, followed by estimating their rigid transformation.
The field has seen substantial progress in recent years with methods that introduce a learning component to solve the task. 

Most learning methods focus on solving the correspondence task~\cite{gojcic2019perfect,bai2020d3feat,Choy2019FCGF,predator}, where a feature extractor is trained to extract point correspondences between two input point clouds. Once the learning step is over, they use the estimated correspondences for computing the rigid pose. 
Due to the low inlier ratio in putative correspondences, these methods strongly rely on hypothesize-and-verify frameworks, \eg~RANSAC~\cite{martin1981random} to compute the pose in a robust manner. 
Recent methods~\cite{qin2022geometric,yew2022regtr} employ advances in the field of transformers to improve the final estimated set of correspondences and remove the dependency on RANSAC, achieving close-to-RANSAC performance. 
However, in these methods too, the objective in the learning process remains to find the best and cleanest matches, ignoring the objective to estimate the rigid pose. 
In addition, they do not achieve end-to-end differentiable training since they still employ robust estimation (\eg,~\cite{predator, qin2022geometric}) combined with the Kabsch-Umeyama algorithm~\cite{yew2022regtr}. 

Other learning-based methods, such as~\cite{wang2019deep,yew2020rpm,wang2019prnet}, directly solve the registration problem by incorporating the pose estimation in their training pipeline. 
Since RANSAC is non-differentiable due to the random sampling, they choose to estimate the alignment using soft correspondences that are computed from local feature similarity scores. 
In contrast to these methods, we employ the aforementioned works on estimating hard correspondences and develop a robust solution to replace RANSAC, that allows for end-to-end differentiable training. 

\begin{figure}[t]
\begin{center}
    \includegraphics[width=0.99\linewidth]{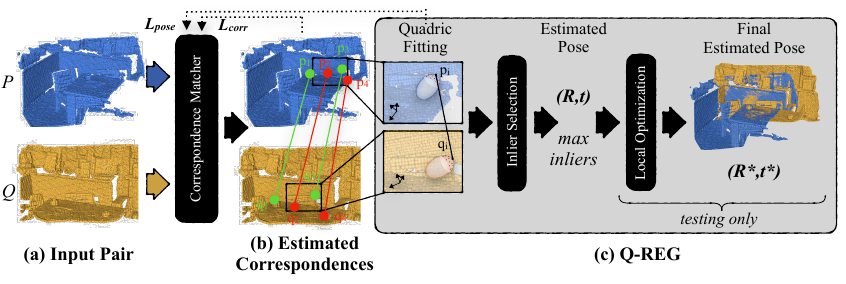}
\end{center}
    \vspace{-5mm}
    \caption{\textbf{\textit{Q-REG} solver.} Given (a) two partially overlapping point clouds as input and (b) the estimated correspondences of a matching method, (c) \textit{Q-REG} leverages the rich local geometry to estimate the rigid pose from a single correspondence, hence enabling end-to-end training of the matcher. \textit{(Best viewed on screen.)}}
\label{fig:teaser}
\end{figure}

In general, RANSAC-like robust estimation is non-differentiable only due to the employed randomized sampling function. Such a sampler is essential to cope with the combinatorics of the problem via selecting random subsets of $m$ correspondences (\eg, $m = 3$ for rigid pose estimation). This allows to progressively explore the $\binom{n}{m}$ possible combinations, where $n$ is the total number of matches.
Actually testing all of them is unbearably expensive in practice, which is what methods like~\cite{qin2022geometric,yew2022regtr} try to avoid. 
This computation bottleneck would be resolved if $m=1$. 
Hence, we design a 1-point solution, \textbf{\textit{Q-REG}}, that utilizes rich geometric cues extracted from local surface patches estimated from observed points (Figure~\ref{fig:teaser}). 
Specifically, we utilize rich geometric information by fitting quadrics (\eg, an ellipsoid) locally to the neighborhoods of an estimated correspondence. 
Moreover, such a solution allows quick outlier rejection by filtering degenerate surfaces and rigid poses inconsistent with motion priors (\eg, to avoid unrealistically large scaling).
\textit{Q-REG} is designed to be deterministic, differentiable, and it replaces RANSAC for point cloud registration. 
It can be used together with any feature-matching or correspondence-matching method.

Since \textit{Q-REG} is fully differentiable, we achieve end-to-end training that optimizes both the correspondence matching and final pose objectives. As such, any learning-based matcher can be extended to being end-to-end trainable. In our experiments, we demonstrate how \textit{Q-REG} consistently improves the performance of state-of-the-art matchers on the \textit{3DMatch}~\cite{zeng20173dmatch}, \textit{KITTI}~\cite{geiger2012we} and \textit{ModelNet}~\cite{wu20153d} datasets. 
It sets new state-of-the-art results on all benchmarks.

\noindent
Our contributions can be summarized as follows:
\begin{itemize}
    \item We develop \textit{Q-REG}, a solution for point cloud registration, estimating the pose from a single correspondence via leveraging local surface patches. 
    It is agnostic to the correspondence matching method. 
    \textit{Q-REG} allows for quick outlier rejection by filtering degenerate solutions and assumption inconsistent motions (\eg, related to scaling).
    \item We extend the above formulation of \textit{Q-REG} to a differentiable setting that allows for end-to-end training of correspondence matching methods with our solver. 
    Thus, we optimize not only over the correspondence matching but also over the final pose.
    \item We demonstrate the effectiveness of \textit{Q-REG} with different baselines on several benchmarks and achieve new state-of-the-art performance across all of them.
\end{itemize}

%% file: sec/2_related_work.tex
\section{Related Work}
\label{sec:Re_work}
\textbf{Correspondence-based Registration Methods.}
The 3D point cloud registration field is well-established and active. Approaches can be grouped into two main categories: \textit{feature-based} and \textit{end-to-end} registration. Feature-based methods comprise two steps: local feature extraction and pose estimation using robust estimators, like RANSAC~\cite{martin1981random}. Traditional methods use hand-crafted features~\cite{tombari2010USC,johnson1999, rusu2008PFH,rusu2009FPFH,tombari2010SHOT} to capture local geometry and, while having good generalization abilities across scenes, they often lack robustness against occlusions. Learned local features have taken over in the past few years, and, instead of using heuristics, they rely on deep models and metric learning~\cite{hermans2017defense,sun2020circle} to extract dataset-specific discriminative local descriptors. These learned descriptors can be divided into patch-based and fully convolutional methods depending on the input. Patch-based ones~\cite{gojcic20193DSmoothNet,ao2021spinnet} treat each point independently, while fully convolutional methods~\cite{Choy2019FCGF,bai2020d3feat,predator} extract all local descriptors simultaneously for the whole scene in a single forward pass.

\textbf{Direct Registration Methods.} 
Recently, end-to-end methods have appeared replacing RANSAC with a differentiable optimization algorithm that targets to incorporate direct supervision from ground truth poses. The majority of these methods~\cite{wang2019deep,wang2019prnet,yew2020rpm} use a weighted Kabsch solver~\cite{kabsch} for pose estimation. Deep Closest Point (DCP)~\cite{wang2019deep} iteratively computes soft correspondences based on features extracted by a dynamic graph convolutional neural network~\cite{wang2019dynamic}, which are then used in the Kabsch algorithm to estimate the transformation parameters. To handle partially overlapping point clouds, methods relax the one-to-one correspondence constraint with keypoint detection~\cite{wang2019prnet} or optimal transport~\cite{yew2020rpm,fischer2021stickypillars}. 
Another line of work replaces local with global feature vectors that are used to regress the pose. PointNetLK~\cite{aoki2019pointnetlk} registers point clouds by minimizing the distance of their latent vectors, in an iterative fashion that resembles the Lucas-Kanade algorithm~\cite{lucas1981iterative}. In~\cite{xu2021omnet}, an approach is proposed for rejecting non-overlapping regions via masking the global feature vector. However, due to the weak feature extractors, there is still a large performance gap compared to hard matching methods.
These direct registration methods primarily work on  synthetic shape datasets~\cite{wu20153d} and often fail in large-scale scenes~\cite{predator}. \textit{Q-REG} uses \textit{hard} correspondences while still being differentiable, via introducing an additional loss component that minimizes the pose error. In addition, as demonstrated in Sec.~\ref{sec:exp}, it works for both real-world indoor~\cite{zeng20173dmatch} and outdoor~\cite{geiger2012we} scene large-scale point clouds and synthetic object-level datasets~\cite{wu20153d}, by setting a new state-of-the-art.

\textbf{Learned Robust Estimators.} 
To address the fact that RANSAC is non-differentiable, other methods either modify it~\cite{brachmann2017dsac} or learn to filter outliers followed by a hypothesize-and-verify framework~\cite{bai2021pointdsc} or a weighted Kabsch optimization~\cite{choy2020deep,pais20203dregnet,gojcic2020learning}. In the latter case, outliers are filtered by a dedicated network, which infers the correspondence weights to be used in the weighted Kabsch algorithm. 
Similarly, we employ the correspondence confidence predicted by a feature extraction network (\eg, by~\cite{predator,qin2022geometric,yew2022regtr}) as weights in the pose-induced loss. 

%% file: sec/3_method.tex
\section{Point Cloud Registration with Quadrics}
\label{cha:chapter4}

\begin{figure*}[h]
\begin{center}
   \includegraphics[width=0.99\linewidth]{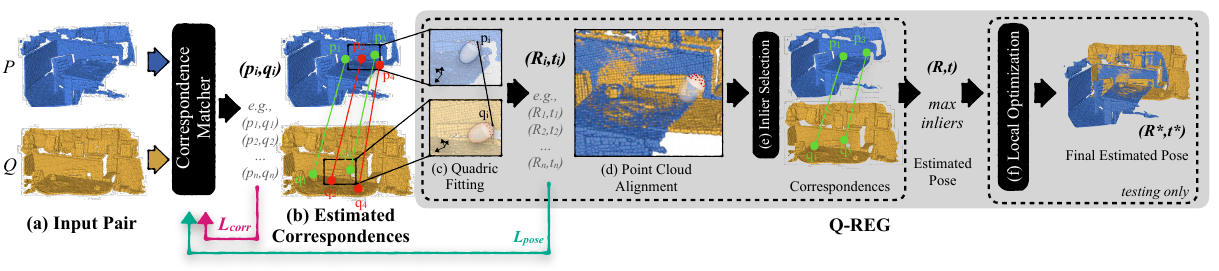}
\end{center}
    \caption[Overview of \textit{Q-REG}]{\textbf{Overview of \textit{Q-REG}.} During inference, given (a) an input pair of partially overlapping point clouds and (b) the output of a correspondence matcher, we (c) perform quadric fitting for each estimated correspondence from which (d) we estimate the rigid pose and (e) compute the inliers given this pose. We iterate over all estimated correspondences, and choose the estimated pose that yields the most inliers. We further improve the result with (f) the local optimization and output the final estimated pose. 
    During training, we back-propagate the gradients to the correspondence matcher and, in addition to its standard loss formulation, we minimize the proposed loss ($L_\text{pose}$) based on the single-correspondence pose estimation. \textit{(Best viewed on screen.)}}
\label{fig:pipeline}
\end{figure*}

We first describe the definition of the point cloud registration problem 
followed by ways of extracting local surface patches that can be exploited for point cloud registration.

\textbf{Problem Definition.} Suppose that we are given two 3D point clouds 
$\mathcal{P} = \{\mathbf{p}_i \in \mathbb{R}^3\ |\ i = 1, ...,\mathnormal{N}\}$ and 
$\mathcal{Q}=\{\mathbf{q}_i \in \mathbb{R}^3\ |\ i = 1, ...,\mathnormal{M}\}$, 
and a set of 3D-3D point correspondences $\mathcal{C}=\{({p}_i,{q}_i)\ |\ {p}_i \in  \mathcal{P}, {q}_i \in  \mathcal{Q},\ i \in [1, K] \}$ extracted, \eg, by the state-of-the-art matchers \cite{predator,qin2022geometric,yew2022regtr}.
The objective is to estimate rigid transformation $\mathbf{T} = \{\mathbf{R},\mathbf{t} \}$ that aligns the point clouds as follows:
\begin{equation}
    \label{eq:f6}
    \min_{\mathbf{R}, \mathbf{t}} \sum\nolimits_{(\mathbf{p}_{x}^*, \mathbf{q}_{y}^*) \in \mathcal{C}^*} \Vert \mathbf{R} \mathbf{p}_{x}^* + \mathbf{t}-\mathbf{q}_{y}^* \Vert_2^2,
\end{equation}
where $\mathbf{R} \in \text{SO}(3)$ is a 3D rotation and $\mathbf{t} \in \mathbb{R}^3$ is a translation vector, and $\mathcal{C}^*$ is the set of ground truth correspondences between $\mathcal{P}$ and $\mathcal{Q}$. 
In practice,  we use putative correspondences instead of ground truth matches, and the set of correspondences often contains a large number of incorrect matches, \ie, outliers. 
Therefore, the objective is formulated as follows:
\begin{equation}
    \label{eq:robust_problem}
    \min_{\mathbf{R}, \mathbf{t}} \sum\nolimits_{(\mathbf{p}_{x}, \mathbf{q}_{y}) \in \mathcal{C}} \rho(\Vert \mathbf{R} \mathbf{p}_{x} + \mathbf{t}-\mathbf{q}_{y} \Vert_2^2),
\end{equation}
where $\rho: \mathbb{R} \to \mathbb{R}$ is a robust loss, \eg, Huber loss. 
The problem is solved by a RANSAC-like~\cite{martin1981random} hypothesize-and-verify framework combined with the Kabsch-Umeyama algorithm~\cite{kabsch}.
We argue in the next sections that, when employing higher-order geometric information, RANSAC can be replaced by exhaustive search improving both the performance and run-time. Figure~\ref{fig:pipeline} illustrates the developed approach, called \textit{Q-REG}.

\subsection{Local Surface Patches}
\label{sec:local_surf_patches}

The main goal in this section is to determine a pair of local coordinate systems $(\mathbf{R}_\Point, \mathbf{R}_\mathbf{q})$ for each correspondence $(\Point, \mathbf{q}) \in \mathcal{C}$, where $\mathbf{R}_\Point$, $\mathbf{R}_\mathbf{q} \in \text{SO}(3)$.
These coordinate systems will be then used to determine rotation $\mathbf{R}$ between the point clouds as $\mathbf{R} = \mathbf{R}_\mathbf{q} \mathbf{R}_\Point^\text{T}$. 
We will describe the method for calculating $\mathbf{R}_\Point$, which is the same for $\mathbf{R}_\mathbf{q}$.
Note that determining translation $\mathbf{t}$ is straightforward as $\mathbf{t} = \mathbf{q} - \Point$.

Suppose that we are given a point $\Point \in \mathcal{P}$ and its $k$-nearest-neighbors $\mathcal{N} \subseteq \mathcal{P}$ such that there exists a correspondence $(\Point, \mathbf{q}) \in \mathcal{C}$, $k \in \mathbb{N}^+$.
One possible solution is to fit a general quadratic surface to the given point and the points in $\mathcal{N}$ and find the principal directions via the first and second-order derivatives at point $\Point$.
These directions can give us a local coordinate system that is determined by the translation and rotation of the local surface in the canonical coordinate system.
Even though this algorithm is widely used in practice, it can suffer from degenerate cases and slow computation time. To address these limitations, we develop the following approach.

The approach which we adopt in this paper is based on fitting a local quadric, \eg ellipsoid, to the point $\Point$ and the points in its neighborhood $\mathcal{N}$.
The general constraint that a 3D quadric surface imposes on a 3D homogeneous point $\hat{\Point}^\text{T}=(x,y,z,1) \in \mathcal{N}$ lying on the surface is
\begin{equation}
    \label{eq:quadric_constraint_1}
    \hat{\Point}^\text{T} \mathbf{Q} \hat{\Point} = 0,
\end{equation}
where $\mathbf{Q}$ is the quadric parameters in matrix form \cite{newman1993model} as:
\begin{equation}
    \label{eq:f36}
    \mathbf{Q} = \begin{pmatrix}
    A & D & E & G \\
    D & B & F & H \\
    E & F & C & I \\
    G & H & I & J \\
    \end{pmatrix}.
\end{equation}
We can rewrite constraint \eqref{eq:quadric_constraint_1} into the form ${\textbf{k}^\text{T}} \textbf{w} = d$, where
\begin{eqnarray*}
    \mathbf{k}^\text{T} & = & (x^2+y^2-2z^2, x^2+z^2-2y^2, 2xy, \\
    & & 2xz, 2yz, 2x, 2y, 2z, 1) \notag, \\
    \mathbf{w}^\text{T} &  = &(A^{\prime}, B^{\prime}, D, E, F, G, H, I, J), \\
     d &=& x^2 + y^2 + z^2, \\
    A^{\prime} &  =& \frac{2A+B}{3} + 1, \\
    B^{\prime} & = &\frac{A-B}{3}.
\end{eqnarray*}
By imposing the constraints on all points, we have:
\begin{equation}
    \label{eq:f44}
    \sum_{i=1}^{|\mathcal{N}|} \mathbf{k}_i {\mathbf{k}_i}^\text{T} \mathbf{w} =\sum_{i=1}^{|\mathcal{N}|} \mathbf{k}_i d_i.
\end{equation}
$|\mathcal{N}|$ is the number of neighbors to which the quadric is fitted (set to 50 in our experiments);  
$d_{i}$ is the squared dist.\ of the $i$th neighbor from the origin.
By solving the above linear equation, we get the coefficients of the quadric surface $\mathbf{Q}$.

As we are interested in finding $\mathbf{Q}$ such that the observed point $\Point$ lies on its surface, we express coefficient $J$ (\ie, the offset from the origin) in Eq.~\ref{eq:f36} with formula
\begin{equation}
    \label{eq:quadric_constraint}
    \mathbf{p}^\text{T} \mathbf{Q} \mathbf{p} = 0.
\end{equation}
Thus, $J$ will not be estimated from the neighborhood but it is preset to ensure that the quadric lies on point $\Point$.

In order to find a local coordinate system, we calculate the new coefficient matrix $\mathbf{P}$ as follows:
\begin{equation}
    \label{eq:P/J}
    \mathbf{P} = \frac{1}{J}\begin{pmatrix}
    A & D & E  \\
    D & B & F  \\
    E & F & C  \\
    \end{pmatrix}.
\end{equation}
Matrix $\mathbf{P}$ can be decomposed by Eigen-decomposition into $\mathbf{P}=\mathbf{V}\boldsymbol{\Sigma}\mathbf{V}^\text{T}$, where $\mathbf{V}=(\mathbf{v}_1,\mathbf{v}_2,\mathbf{v}_3)$, projecting the fitted points to canonical coordinates, and $\boldsymbol{\Sigma}=\text{diag}(\lambda_1,\lambda_2,\lambda_3)$ comprises the eigenvalues.

Matrix $\mathbf{V}$ contains the three main axes that map the quadric, fitted to point $\Point$ and its local neighborhood $\mathcal{N}$, to canonical form. 
It is easy to see that, due to its local nature, the local surface is invariant to the rigid translation and rotation of the point cloud. 
Thus, it is a repeatable feature under rotations and translations of the underlying 3D point cloud. $\boldsymbol{\Sigma}$ contains the three eigenvalues that are proportional to the reciprocals of the lengths $l_1$, $l_2$, $l_3$ of three axes squared.

\subsection{Rigid Transformation from Surface Matches}
\label{sec:solver}

Suppose that we are given sets of local coordinate systems $\mathcal{V}^\mathcal{P}$ and $\mathcal{V}^\mathcal{Q}$ associated with points on the found 3D-3D point correspondences, estimated as explained in the previous section.
Given correspondence $(\Point, \mathbf{q}) \in \mathcal{C}$, we know the local coordinates systems $\mathbf{V}^\mathcal{P}_\Point \in \mathcal{V}^\mathcal{P}$ and $\mathbf{V}^\mathcal{Q}_\mathbf{q} \in \mathcal{V}^\mathcal{Q}$ at, respectively, points $\Point$ and $\mathbf{q}$.
Due to the local surfaces being translation and rotation invariant,
the coordinate systems must preserve the rigid transformation applied to the entire point cloud. 
Thus, $\mathbf{R} =  \mathbf{V}^\mathcal{Q}_\mathbf{q} \mathbf{P} (\mathbf{V}^\mathcal{P}_\Point)^\text{T} \in \text{SO}(3)$ is the rotation between the point clouds, where $\mathbf{P}$ is an unknown permutation matrix assigning the axes in the first coordinate system to the axes in the second one. 

There are three cases that we have to account for. 
Ideally, the lengths of the three axes $\mathbf{L}^a = (l_1^a, l_2^a, l_3^a)^\text{T}$ have a distinct ordering such that $l_1^a > l_2^a > l_3^a$, $a \in \{\mathcal{P}, \mathcal{Q}\}$.
In this case, the permutation matrix can be determined such that it assigns the longest axis in $\mathbf{V}^\mathcal{Q}_\mathbf{q}$ to the longest one in $\mathbf{V}^\mathcal{P}_\Point$, and so on. 
This procedure builds on the assumption that there is no or negligible anisotropic scaling in the point clouds and thus, the relative axis lengths remain unchanged.
Also, having this assignment allows us to do the matching in a scale-invariant manner while enabling us to calculate a uniform scaling of the point clouds -- or to early reject incorrect matches that imply unrealistic scaling.
In this case, the problem is solved from a single correspondence.

The second case is when two axes have the same lengths, \eg, $l_1^a \approx l_2^a$, and $l_3^a$ is either shorter or longer than them. 
In this scenario, only $l_3^a$ can be matched between the point clouds. 
This case is equivalent to having a corresponding oriented point pair.
It gives us an additional constraint for estimating the rotation matrix.
However, the rotation around axis $l_3^a$ is unknown and has to be estimated from another point correspondence.
While this is a useful solution to reduce the required number of points from three to two, it does not allow solving from a single correspondence.  

In the third case, when $l_1^a \approx l_2^a \approx l_3^a$, we basically are given a pair of corresponding spheres that provide no extra constraints on the unknown rotation. 

In the proposed algorithm, we keep only those correspondences from $\mathcal{C}$ where the local surface patches are of the first type, \ie, they lead to enough constraints to estimate the rigid transformation from a single correspondence. 
Specifically, we keep only those correspondences, where $l_1^a \neq l_2^a \neq l_3^a$ with $10^{-3}$ tolerance.
Next, we will discuss how this approach can be used for training 3D-3D correspondence matching algorithms with robust estimation in an end-to-end manner.

\subsection{End-to-End Training}
\label{sec:training}

Benefiting from the rich information extracted via local surfaces (described in the prev.\ section), the presented solver estimates the rigid pose from a single 3D quadric match. 
This unlocks end-to-end differentiability, where the gradients of the matcher network can be propagated through the robust estimator to a loss directly measuring the pose error per correspondence.
This enables using test-time evaluation metrics to optimize the end-to-end training.

\textbf{Loss.} In order to calculate a pose-induced loss from each correspondence, we first fit quadrics to local neighborhoods.
This step has to be done only once, prior to the loss calculation, as the point clouds do not change. 
Suppose that we are given a set of correspondences $\mathcal{C}=\{(\Point, \mathbf{q}, \mathbf{V}_\Point^\mathcal{P}, \mathbf{V}_\mathbf{q}^\mathcal{Q}) \; | \; \Point \in  \mathcal{P}, \; \mathbf{q} \in \mathcal{Q}, \; \mathbf{V}_\Point^\mathcal{P} \in \mathcal{V}^\mathcal{P}, \; \mathbf{V}_\mathbf{q}^\mathcal{Q} \in \mathcal{V}^\mathcal{Q}\}$ equipped with their local quadrics and a solver $\phi: \mathcal{P} \times \mathcal{Q} \times \mathcal{V}^\mathcal{P} \times \mathcal{V}^\mathcal{Q} \to \text{SE}(3)$, as described in Sec.~\ref{sec:solver}, we can estimate the rigid transformation $\mathbf T = (\mathbf R, \mathbf t) \in \text{SE}(3)$ from a single correspondence. 
Given a correspondence $(\Point, \mathbf{q}, \mathbf{V}_\Point^\mathcal{P}, \mathbf{V}_\mathbf{q}^\mathcal{Q})$ and the pose estimated from it $\mathbf T_{\mathbf p, \mathbf q} = \phi(\mathbf{p}, \mathbf{q}, \mathbf{V}_\Point^\mathcal{P}, \mathbf{V}_\mathbf{q}^\mathcal{Q})$,
The error is formalized as follows:
\begin{equation}
    \label{eq:f78}
    \epsilon(\mathbf T_{\mathbf p, \mathbf q}) = \sqrt{\frac{1}{\left|\mathcal{C}\right|}\sum_{\left(\Point_i, \mathbf{q}_i, ... \right) \in \mathcal{C}} \Vert \mathbf T_{\mathbf p, \mathbf q} \mathbf{p}_i - \mathbf{q}_i \Vert_2^2 },
\end{equation}
where the RMSE of the pose is calculated by transforming the correspondences. 
The loss obtained by iterating through all correspondences is as follows:
\begin{equation}
    \label{eq:f79}
    L_\text{pose} = \sum_{(\Point, \mathbf{q}, \mathbf{V}_\Point^\mathcal{P}, \mathbf{V}_\mathbf{q}^\mathcal{Q}) \in \mathcal{C}} \left(1 - \frac{\min(\epsilon(\mathbf T_{\Point,\mathbf{q}}), \gamma)}{\gamma}-s \right),
\end{equation}
where $\gamma \in \mathbb{R}$ is a threshold and $s$ is the score of the point correspondence predicted by the matching network. 
The proposed $L_\text{pose}$ can be combined with any of the widely used loss functions, \eg, registration loss. 
It bridges the gap between correspondence matching and registration and unlocks the end-to-end training.

\textbf{Inference Time.} While the proposed \textit{Q-REG} propagates the gradients at training time, during inference, we equip it with components that ensure high accuracy but are non-differentiable.
\textit{Q-REG} iterates through the poses calculated from all tentative correspondences, by the proposed single-match solver, in an exhaustive manner. 
For each match, the pose quality is calculated as the cardinality of its support, \ie, the number of inliers. 
After the best model is found, we apply local optimization similar to~\cite{lebeda2012fixing}, a local re-sampling and re-fitting of inlier correspondences based on their normals (coming from the fitted quadrics) and positions.

%% file: sec/4_experiments.tex
\section{Experiments}
\label{sec:exp}

We evaluate \textit{Q-REG} with three state-of-the-art matchers (Predator~\cite{predator}, RegTR~\cite{yew2022regtr}, and GeoTr~\cite{qin2022geometric}) on real, indoor point cloud datasets \textit{3DMatch}~\cite{zeng20173dmatch} and \textit{3DLoMatch}~\cite{predator}.
We also evaluate \textit{Q-REG} with Predator and GeoTr on the real, outdoor dataset \textit{KITTI}~\cite{geiger2012we} and on the synthetic, object-centric datasets \textit{ModelNet}~\cite{wu20153d} and \textit{ModelLoNet}~\cite{predator}.   
We compare \textit{Q-REG} with other estimators that predict rigid pose from correspondences on the \textit{3DLoMatch} datasets.
Furthermore, we evaluate the importance of different \textit{Q-REG} components on the best-performing matcher on \textit{3DLoMatch}, as well as run-time during inference.  

\vspace{1mm} \noindent \textbf{3DMatch \& 3DLoMatch.}
\label{sec:3dmatch_3dlomatch}
The \textit{3DMatch}~\cite{zeng20173dmatch} dataset contains 62 scenes in total, with 46 used for training, 8 for validation, and 8 for testing. We use the training data preprocessed by Huang et al.~\cite{predator} and evaluate on both \textit{3DMatch} and \textit{3DLoMatch}~\cite{predator} protocols. The point cloud pairs in \textit{3DMatch} have more than 30\% overlap, whereas those in \textit{3DLoMatch} have a low overlap of 10\% - 30\%. Following prior work~\cite{qin2022geometric,yew2022regtr}, we evaluate the following metrics: (i) Registration Recall (RR), which measures the fraction of successfully registered pairs, defined as having a correspondence RMSE below 0.2 m; (ii) Relative Rotation Error (RRE); and (iii) Relative Translation Error (RTE). Both (ii) and (iii) measure the accuracy of successful registrations. Additionally, we report the mean RRE, RTE, and RMSE. In this setting, we evaluate over all valid pairs\footnote{According to~\cite{zeng20173dmatch}, a valid pair is a pair of non-consecutive frames.} instead of only those with an RMSE below 0.2 m, and we provide a simple average over all valid pairs instead of the median value of each scene followed by the average over all scenes. These metrics will show how consistently well (or not) a method performs in registering scenes. 

We report several learned correspondence-based algorithms on the two datasets. For~\cite{gojcic2019perfect,Choy2019FCGF,bai2020d3feat}, we tabulate the results as reported in their original papers. For~\cite{predator,yew2022regtr,qin2022geometric}, we evaluate them with and without the \textit{Q-REG} solver on all metrics. We also report methods that do not employ RANSAC~\cite{xu2021omnet,choy2020deep,cao2021pcam} -- results are taken from~\cite{yew2022regtr}. 

The results for \textit{3DLoMatch} and \textit{3DMatch} are tabulated in Tables~\ref{tab:3dlomatch} and \ref{tab:3dmatch} respectively. Note that, unless differently stated, hereafter the best values per group are in \textbf{bold} and the absolute best is \underline{\textbf{underlined}}. Also, \textit{Q-REG} means that the solver is used only in inference and \textbf{\textit{Q-REG*}} means it is used in both end-to-end training and inference. In the latter case, we train from scratch the correspondence matching network with the addition of the pose-induced loss.
We use the default training procedure and parameters specified for each particular matcher when retraining. 
$50K$ refers to the RANSAC iterations. Last, if nothing is added next to a method, the standard formulation is used.

In all three matchers, incorporating \textit{Q-REG} in inference time yields an increase in RR that ranges from 1.0 to 6.2\% in \textit{3DLoMatch} and from 0.9 to 1.6\% in \textit{3DMatch}. The range difference between the two datasets is expected, since \textit{3DMatch} is more saturated and the gap for improvement is small. Using \textit{Q-REG} for inference achieves the second-best results overall (GeoTr + Q-REG). Even in the case of RegTR, where the putative correspondence set is smaller than other two methods and applying RANSAC ends in decreasing performance~\cite{yew2022regtr}, \textit{Q-REG} can still provide a boost in all metrics. When training end-to-end the best-performing matcher, GeoTr, we gain further boost and achieve the best results overall in both datasets, setting a new benchmark (GeoTr + \textbf{\textit{Q-REG*}}). We observe this behavior not only on the standard metrics (RR, RRE, RTE), but also at the Mean RRE, RTE, and RMSE. As expected, \textit{Q-REG} results in smaller errors regardless of the matcher. 
Additional results of Inlier Ratio (IR) and Feature Matching Recall (FMR) can be found in the supplementary material.

\begin{table}[t]
    \footnotesize    
    \begin{center}
    \resizebox{\columnwidth}{!}{
    \begin{tabular}{l|ccc|ccc}
        \toprule
        \multirow{2}{*}{Model} & RR & RRE & RTE & \multicolumn{3}{c}{\textit{Mean}} \\
         & ($\%$)$\uparrow$ & ($^{\circ}$)$\downarrow$ & (cm)$\downarrow$ & RRE $\downarrow$ & RTE $\downarrow$ & RMSE (cm)$\downarrow$ \\
	    \midrule
	    3DSN~\cite{gojcic2019perfect}   & 33.0 & 3.53 & 10.3 & - & - & - \\ 
	    FCGF~\cite{Choy2019FCGF}        & 40.1 & \textbf{3.15} & 10.0 & - & - & - \\
	    D3Feat~\cite{bai2020d3feat}     & 37.2 & 3.36 & 10.3 & - & - & - \\
	    OMNet~\cite{xu2021omnet}        & 8.4  & 7.30 & 15.1 & - & - & - \\
	    DGR~\cite{choy2020deep}         & 48.7 & 3.95 & 11.3 & - & - & - \\
	    PCAM~\cite{cao2021pcam}         & \textbf{54.9} & 3.53 & \textbf{9.9}  & - & - & - \\
	    \midrule
	    Predator~\cite{predator} + 50k            & 60.4 & 3.21 & 9.5  & 30.07 & 76.2 & 72.8 \\
	    Predator~\cite{predator} + Q-REG & \textbf{66.6} & \textbf{2.70} & \textbf{8.1}  & \textbf{28.44} & \textbf{71.9} & \textbf{68.8}\\
	    \midrule
	    RegTR~\cite{yew2022regtr}                  & 64.8 & 2.83 & 8.0 & 23.05 & 64.4 & 55.8 \\
        RegTR~\cite{yew2022regtr} + 50K     & 64.3 & 2.92 & 8.5 & 21.90 & 62.5 & 54.5 \\
        RegTR~\cite{yew2022regtr} + Q-REG & \textbf{65.3} & \textbf{2.81} & \textbf{7.8} & \textbf{21.43} & \textbf{60.9} & \textbf{53.3}\\
        \midrule
	    GeoTr~\cite{qin2022geometric}                            & 74.1 & 2.59 & {7.3} & 23.15 & 58.3 & 57.8 \\
	    GeoTr~\cite{qin2022geometric} + 50K                      & 75.0 & 2.54 & 7.7 & 22.69 & 57.8 & 57.3 \\
	    GeoTr~\cite{qin2022geometric} + Q-REG           & {77.1} & {2.44} & 7.7 & 16.70 & \underline{\textbf{46.0}} & 44.6 \\
	    GeoTr~\cite{qin2022geometric} + \textbf{\textit{Q-REG*}} & \underline{\textbf{78.3}} & \underline{\textbf{2.38}} & \underline{\textbf{7.2}} & \underline{\textbf{15.65}} & {46.3} & \underline{\textbf{42.5}} \\
	    
    	\bottomrule
    \end{tabular}
    }
    \end{center}
    \vspace{-5mm}
\caption{Evaluation of state-of-the-art matchers on the \textbf{\textit{3DLoMatch}}~\cite{predator} dataset. The best values are \textbf{bold} in each group. The absolute best are \underline{\textbf{underlined}}. 
}
\label{tab:3dlomatch}
\end{table}

\begin{table}[t]
    \footnotesize
    \begin{center}
    \resizebox{\columnwidth}{!}{
    \centering
    \begin{tabular}{l|ccc|ccc}
        \toprule
        \multirow{2}{*}{Model} & RR & RRE & RTE & \multicolumn{3}{c}{\textit{Mean}} \\
         & ($\%$)$\uparrow$ & ($^{\circ}$)$\downarrow$ & (cm)$\downarrow$ & RRE $\downarrow$ & RTE $\downarrow$ & RMSE (cm)$\downarrow$ \\
	    \midrule
	    3DSN~\cite{gojcic2019perfect}  & 78.4 & 2.20 & 7.1 & - & - & - \\ 
	    FCGF~\cite{Choy2019FCGF}       & 85.1 & 1.95 & 6.6 & - & - & - \\
	    D3Feat~\cite{bai2020d3feat}    & 81.6 & 2.16 & 6.7 & - & - & - \\
	    OMNet~\cite{xu2021omnet}       & 35.9 & 4.17 & 10.5 & - & - & - \\
	    DGR~\cite{choy2020deep}        & 85.3 & 2.10 & 6.7 & - & - & - \\
	    PCAM~\cite{cao2021pcam}        & \textbf{85.5} & \textbf{1.81} & \textbf{5.9} & - & - & - \\
	    \midrule
	    Predator~\cite{predator} + 50k              & 89.3 & 1.98 & 6.5 & 6.80 & 20.2 & 18.3 \\
	    Predator~\cite{predator} + Q-REG   & \textbf{90.6} & \textbf{1.74} & \textbf{5.7} & \textbf{6.78} & \textbf{20.0} & \textbf{18.1} \\
	    \midrule
	    RegTR~\cite{yew2022regtr}                  & 92.0 & \textbf{1.57} & \textbf{\underline{4.9}} & 5.31 & 17.0 & 13.8 \\
        RegTR~\cite{yew2022regtr} + 50K     & 91.3 & 1.72 & 5.9 & 5.26 & 17.5 & 14.7 \\
        RegTR~\cite{yew2022regtr} + Q-REG          & \textbf{92.2} & \textbf{1.57} & \textbf{\underline{4.9}} & \textbf{5.13} & \textbf{16.5} & \textbf{13.6} \\
        \midrule
	    GeoTr~\cite{qin2022geometric}                            & 92.5 & {1.54} & \textbf{5.1} & 7.04 & 19.4 & 17.6 \\
	    GeoTr~\cite{qin2022geometric} + 50K                      & 92.2 & 1.66 & 5.6 & 6.85 & 18.7 & 17.1 \\
	    GeoTr~\cite{qin2022geometric} + Q-REG                    & 93.8 & 1.57 & {5.3} & {4.74} & {15.0} & {12.8} \\
	    GeoTr~\cite{qin2022geometric} + \textbf{\textit{Q-REG*}} & \textbf{\underline{95.2}} &  \textbf{\underline{1.53}} & {5.3} & \textbf{\underline{3.70}} & \textbf{\underline{12.5}} & \textbf{\underline{10.7}} \\
	    
    	\bottomrule
    \end{tabular}
    }
    \end{center}
\vspace{-5mm}
\caption{Evaluation of state-of-the-art matchers on the \textbf{\textit{3DMatch}}~\cite{zeng20173dmatch} dataset. The best values are \textbf{bold} in each group. The absolute best are \underline{\textbf{underlined}}.
}
\label{tab:3dmatch}
\end{table}

\begin{figure*}[t]
\begin{center}
   \includegraphics[width=0.99\linewidth]{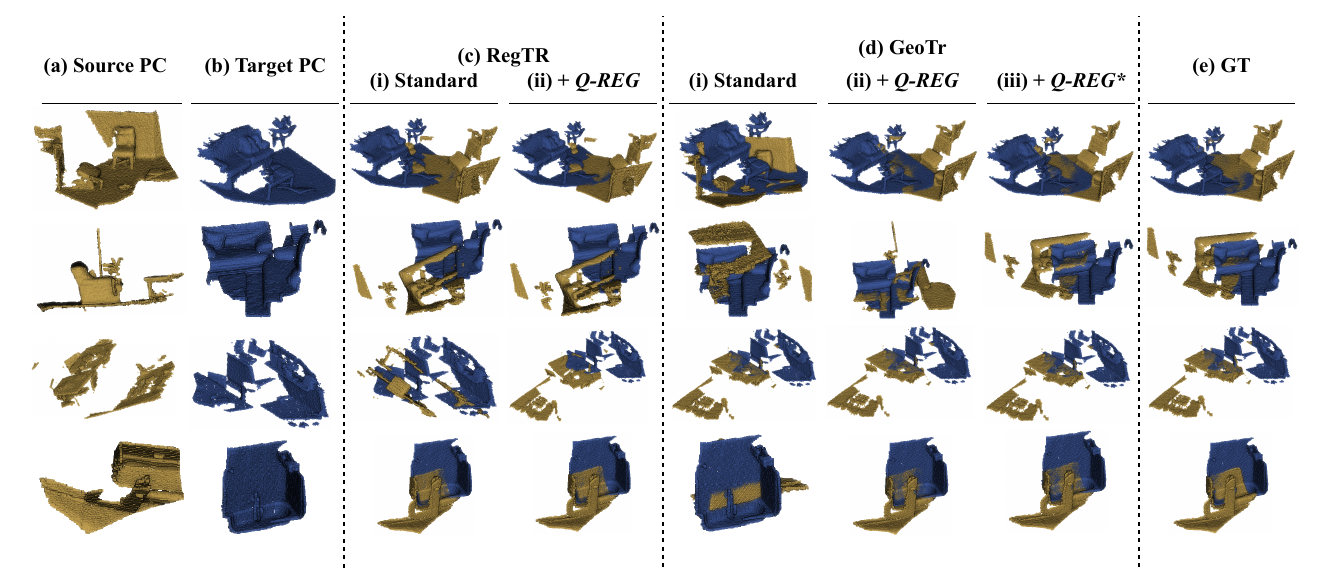}
\end{center}
\vspace{-5mm}
    \caption[Qualitative Results]{\textbf{Qualitative Results.} We showcase registration examples of RegTR~\cite{yew2022regtr} and GeoTr~\cite{qin2022geometric} with and without \textit{Q-REG} for the \textit{3DLoMatch} (first and third rows) and \textit{3DMatch} (second and fourth rows) datasets. \textit{(Best viewed on screen.)}}
\label{fig:quals}
\end{figure*}

Qualitative results are shown in Figure~\ref{fig:quals}. In the first and second row, GeoTr+\textit{\textbf{Q-REG*}} achieves a good alignment of the point clouds when all other methods fail.
This means that using \textit{Q-REG} in end-to-end training can provide additional improvements in performance by learning how to better match correspondences together with the objective of rigid pose estimation, and not in isolation as it happens in all other cases.
In the third row, the standard formulation already produces well-aligned point clouds, and the addition of \textit{Q-REG} slightly refines this output. However, in the case of RegTR, we can see the most improvement. The standard formulation fails to achieve a good alignment and \textit{Q-REG} is able to recover a good pose. This means that our method is able to identify robust correspondences and remove spurious ones.
When RegTR finds a correct pose, \textit{Q-REG} can further optimize it, as shown in the fourth row. In the same example, although GeoTR fails to infer a good pose, both \textit{Q-REG} and \textit{\textbf{Q-REG*}} are able to recover it. Additional qualitative results and  
plots can be found in the supp.\ material.

\vspace{1mm} \noindent \textbf{KITTI.}
\label{sec:KITTI}
The \textit{KITTI} odometry dataset~\cite{geiger2012we} contains 11 sequences of LiDAR-scanned outdoor driving scenarios. We follow~\cite{bai2020d3feat, Choy2019FCGF, predator, qin2022geometric} and split it into train/val/test sets as follows: sequences 0-5 for training, 6-7 for validation and 8-10 for testing. As in~\cite{bai2020d3feat, Choy2019FCGF, predator, qin2022geometric}, we refine the provided ground truth poses using ICP~\cite{besl1992method} and only use point cloud pairs that are captured within 10m range of each other. 
Following prior work~\cite{predator, qin2022geometric}, we evaluate the following metrics: (1) Registration Recall (RR), which is the fraction of point cloud pairs with RRE and RTE both below certain thresholds (\ie, RRE$<$5~and RTE$<$2m), (2) Relative Rotation Error (RRE), and (3) Relative Translation Error (RTE).

We report several recent algorithms on the \textit{KITTI} odometry dataset~\cite{geiger2012we}. For~\cite{yew20183dfeat, Choy2019FCGF, bai2020d3feat}, results are taken from~\cite{qin2022geometric}. For~\cite{qin2022geometric, predator}, we evaluate them with and without \textit{Q-REG}, similarly to the Sec.~\ref{sec:3dmatch_3dlomatch}. The results on the \textit{KITTI} dataset are in Table~\ref{tab:KITTI}. Here as well, we observe a similar trend in the results, with \textit{Q-REG} boosting the performance of all matchers. Despite saturation of methods on \textit{KITTI}, using the \textit{Q-REG} solver during inference provides improvements in both RRE and RTE. Predator with \textit{Q-REG} achieves the best results overall on both datasets (Predator + Q-REG). In addition, when \textit{Q-REG} is used for both inference and end-to-end training, the results of GeoTr are also improved with respect to its standard formulation (GeoTr + \textbf{\textit{Q-REG*}}). This indicates that \textit{Q-REG} has similar behavior in point clouds of lower density and different distribution. Additional 
qualitative results and plots are provided in the supp. material.

\begin{table}[t]
    \footnotesize
    \begin{center}
    \resizebox{\columnwidth}{!}{
    \begin{tabular}{l|ccc}
        \toprule
        \multirow{2}{*}{Method} & \multicolumn{3}{c}{KITTI~\cite{geiger2012we}} \\
        & RR ($\%$)$\uparrow$ & RRE ($^{\circ}$)$\downarrow$ & RTE (cm)$\downarrow$ \\
	    \midrule
	    3DFeat-Net~\cite{yew20183dfeat}  & 96.0 & \textbf{0.25} & 25.9 \\
	    FCGF~\cite{Choy2019FCGF}         & 96.0 & 0.30 & 9.5 \\
	    D3Feat~\cite{bai2020d3feat}      & \textbf{\underline{99.8}} & 0.30 & \textbf{7.2} \\
        \midrule
        Predator~\cite{predator} + 50K           & \textbf{\underline{99.8}} & 0.27 & 6.8 \\
        Predator~\cite{predator} + Q-REG   & \textbf{\underline{99.8}} & \textbf{\underline{0.16}} & \textbf{\underline{3.9}}\\
        \midrule
        GeoTr~\cite{qin2022geometric}      & \textbf{\underline{99.8}} & 0.24 & 6.8\\
        GeoTr~\cite{qin2022geometric} + 50K     & \textbf{\underline{99.8}} & 0.26 & 7.5\\
        GeoTr~\cite{qin2022geometric} + Q-REG                    & \textbf{\underline{99.8}} & 0.20 & 6.0\\
	GeoTr~\cite{qin2022geometric} + \textbf{\textit{Q-REG*}} & \textbf{\underline{99.8}} & \textbf{0.18} & \textbf{5.4}\\
    	\bottomrule
    \end{tabular}
    }
    \end{center}
    \vspace{-5mm}
\caption{Evaluation of state-of-the-art matchers on the \textbf{\textit{KITTI}}~\cite{geiger2012we} dataset. The best values are \textbf{bold} in each group. The absolute best are \underline{\textbf{underlined}}.}
\label{tab:KITTI}
\end{table}

\vspace{1mm} \noindent \textbf{ModelNet \& ModelLoNet.}
\label{sec:modelNet}
The \textit{ModelNet}~\cite{wu20153d} dataset contains  12,311 3D CAD models of man-made objects from 40 categories, with 5,112 used for training, 1,202 for validation, and 1,266 for testing. We use the partial scans created by Yew et al.~\cite{yew2020rpm} for evaluating on \textit{ModelNet} and those created by Huang et al.~\cite{predator} for evaluating on \textit{ModelLoNet}. The point cloud pairs in \textit{ModelNet} have 73.5\% overlap on average, whereas those in \textit{ModelLoNet} have 53.6\%. Following prior work~\cite{predator,yew2022regtr}, we evaluate the following metrics: (i) Chamfer Distance (CD) between registered point clouds; (ii) Relative Rotation Error (RRE); and (iii) Relative Translation Error (RTE). We report several recent algorithms on the two datasets. For~\cite{aoki2019pointnetlk,predator}, we tabulate the results as reported in their original papers. For~\cite{wang2019deep,yew2020rpm,xu2021omnet}, results are taken from~\cite{yew2022regtr}. For~\cite{yew2022regtr,qin2022geometric}, we evaluate them with and without \textit{Q-REG}, similarly as before. 

The results for \textit{ModelNet}~\cite{wu20153d} and \textit{ModelLoNet}~\cite{predator} are tabulated in Tables~\ref{tab:modelnet} and \ref{tab:modelLOnet}, respectively. Here as well, we observe a similar trend in the results, with \textit{Q-REG} boosting the performance of all matchers. RegTR with \textit{Q-REG} achieves the best results overall on both datasets (RegTR + Q-REG). 
In addition, both when \textbf{\textit{Q-REG*}} is used for inference and end-to-end training, the results of GeoTr are also improved with respect to its standard formulation.

\begin{table}[t]
    \footnotesize
    \begin{center}
    \resizebox{\columnwidth}{!}{
    \begin{tabular}{l|ccc}
        \toprule
        \multirow{2}{*}{Method} & \multicolumn{3}{c }{ModelNet~\cite{wu20153d}}\\
        & CD $\downarrow$ & RRE ($^{\circ}$)$\downarrow$ & RTE (cm)$\downarrow$\\
	    \midrule
	    PointNetLK~\cite{aoki2019pointnetlk}  & 0.02350 & 29.73 & 29.7 \\
	    OMNet~\cite{xu2021omnet}              & 0.00150 & 2.95 & 3.2  \\
	    DCP-v2~\cite{wang2019deep}            & 0.01170 & 11.98 & 17.1 \\
	    RPM-Net~\cite{yew2020rpm}             & \textbf{0.00085} & \textbf{1.71} & \textbf{1.8}  \\
        Predator~\cite{predator}              & 0.00089 & 1.74 & 1.9  \\ 
        \midrule
        RegTR~\cite{yew2022regtr}                  & 0.00078 & {1.47} & {1.4}  \\
        RegTR~\cite{yew2022regtr} + 50K     & 0.00091 & 1.82 & 1.8 \\
        RegTR~\cite{yew2022regtr} + Q-REG          & \underline{\textbf{0.00074}} & \underline{\textbf{1.35}} & \underline{\textbf{1.3}} \\
        \midrule
        GeoTr~\cite{qin2022geometric}                            & 0.00083 & 2.16 & 2.0\\
	    GeoTr~\cite{qin2022geometric} + 50K                      & 0.00095 & 2.40 & 2.2\\
	    GeoTr~\cite{qin2022geometric} + Q-REG                    & 0.00078 & 1.84 & 1.7 \\
	    GeoTr~\cite{qin2022geometric} + \textbf{\textit{Q-REG*}} & \textbf{0.00076} & \textbf{1.73} & \textbf{1.5}\\
    	\bottomrule
    \end{tabular}
    }
    \end{center}
    \vspace{-5mm}
\caption[Evaluation of state-of-the-art matchers on the \textit{ModelNet} dataset]{Evaluation of state-of-the-art matchers on the \textbf{\textit{ModelNet}}~\cite{wu20153d} dataset. The best values are \textbf{bold} in each group. The absolute best are \underline{\textbf{underlined}}.} 
\label{tab:modelnet}
\end{table}

\begin{table}[t]
    \footnotesize
    \begin{center}
    \resizebox{\columnwidth}{!}{
    \begin{tabular}{l|ccc}
        \toprule
        \multirow{2}{*}{Method} & \multicolumn{3}{c}{ModelLoNet~\cite{predator}} \\
        & CD $\downarrow$ & RRE ($^{\circ}$)$\downarrow$ & RTE (cm)$\downarrow$ \\
	    \midrule
	    PointNetLK~\cite{aoki2019pointnetlk}  & 0.0367 & 48.57 & 50.7 \\
	    OMNet~\cite{xu2021omnet}              & 0.0074 & 6.52 & 12.9 \\
	    DCP-v2~\cite{wang2019deep}            & 0.0268 & 6.50 & 30.0 \\
	    RPM-Net~\cite{yew2020rpm}             & \textbf{0.0050} & 7.34 & \textbf{12.4} \\
        Predator~\cite{predator}              & 0.0083 & \textbf{5.24} & 13.2 \\ 
        \midrule
        RegTR~\cite{yew2022regtr}                  & {0.0037} & 3.93 & 8.7 \\
        RegTR~\cite{yew2022regtr} + 50K     & 0.0039 & 4.23 & 9.2 \\
        RegTR~\cite{yew2022regtr} + Q-REG          & \underline{\textbf{0.0034}} & \underline{\textbf{3.65}} & \textbf{8.1} \\
        \midrule
        GeoTr~\cite{qin2022geometric}                            & 0.0050 & 4.49 & 7.6\\
	    GeoTr~\cite{qin2022geometric} + 50K                      & 0.0050 & 4.27 & 8.0\\
	    GeoTr~\cite{qin2022geometric} + Q-REG                    & 0.0044 & 3.87 & 7.0\\
	    GeoTr~\cite{qin2022geometric} + \textbf{\textit{Q-REG*}} & \textbf{0.0040} & \textbf{3.73} & \underline{\textbf{6.5}}\\
    	\bottomrule
    \end{tabular}
    }
    \end{center}
    \vspace{-5mm}
\caption{Evaluation of state-of-the-art matchers on the \textbf{\textit{ModelLoNet}}~\cite{predator} dataset. The best values are \textbf{bold} in each group. The absolute best are \underline{\textbf{underlined}}.}
\label{tab:modelLOnet}
\end{table}

\subsection{Comparison with Other Estimators}
\label{sec:comp_est}

We compare \textit{Q-REG} with other estimators that predict rigid pose from correspondences on the \textit{3DLoMatch} dataset, using the state-of-the-art matcher GeoTr~\cite{qin2022geometric} as the correspondence extractor (best performing on this dataset). We evaluate the following estimators: (i) \textbf{GeoTr + WK}: weighted variant of the Kabsch-Umeyama algorithm~\cite{umeyama1991least}. (ii) \textbf{GeoTr + ICP}: Iterative closest point (ICP)~\cite{besl1992method} initialized with 50K RANSAC. 
(iii) \textbf{GeoTr + PointDSC}: PointDSC~\cite{bai2021pointdsc} with their pre-trained model from ~\cite{PointDSC_Github}.
(iv) \textbf{GeoTr + SC2-PCR}: GeoTr with SC2-PCR~\cite{chen2022sc2}.
(v) \textbf{GeoTr + Q-REG w/ PCA}: PCA instead of our quadric fitting to determine the local coordinate system. 
(vi) \textbf{GeoTr + Q-REG w/ PCD}: Use principal direction as explained in Sec.~\ref{sec:local_surf_patches}. 
(vii) \textbf{GeoTr + Q-REG}: Our quadric-fitting solver is used only in inference. 
(viii) \textbf{GeoTr + \textit{Q-REG*}}: Our solver is used in both end-to-end training and inference. 

The results are in Table~\ref{tab:tbd_3dlomatch} (\textit{best results in \textbf{bold}, second best are \underline{underlined}}). Among all methods, \textbf{\textit{Q-REG*}} performs the best in the majority of the metrics. 
GeoTr + WK shows a large performance gap with other methods since it utilizes soft correspondences and is not robust enough to outliers. 
GeoTr + ICP relies heavily on the initialization and, thus, often fails to converge to a good solution.
GeoTr + PointDSC and GeoTr + SC2-PCR have comparable results as LGR but are noticeably worse than \textit{Q-REG}.
GeoTr + Q-REG w/ PCA leads to less accurate results than w/ quadric fitting, which demonstrates the superiority of \textit{Q-REG} that utilizes quadrics.
GeoTr + Q-REG w/ PCD shows comparable performance as our \textit{Q-REG}. This is expected since both methods have the same geometric meaning. However, there is a substantial difference in runtime (secs): 0.166 for quadric fitting versus 0.246 for PCD. Similar comparison results on the \textit{3DMatch dataset} is in the supplementary material.

\subsection{Ablation Studies}
\label{sec:ablation}

We evaluate the contribution of each component in the \textit{Q-REG} solver to the best-performing matcher on the \textit{3DLoMatch} dataset, the state-of-the-art GeoTr~\cite{qin2022geometric}. We evaluate the following self-baselines (i, ii and iii are inference only): 
(i) \textbf{GeoTr + Q}: Our quadric-fitting 1-point solver. 
(ii) \textbf{GeoTr + QL}: We extend the quadric fitting with local optimization (LO) as discussed in Sec.~\ref{sec:training}.
(iii) \textbf{GeoTr + RL}: We replace quadric fitting with RANSAC 50K in (ii).
(iv) \textbf{GeoTr + QT}: Our quadric-fitting solver is used in end-to-end training -- during inference we do not employ LO; and (v) \textbf{GeoTr + QTL}: Our quadric-fitting 1-point solver is used in end-to-end training followed by inference with LO.

The results are reported in Table~\ref{tab:abl_3dlomatch} (\textit{best results in \textbf{bold}, second best are \underline{underlined}}). 
\textbf{\textit{Q-REG*}} performs the best in the majority of the metrics. 
Specifically, we observe that there is a substantial increase in RR by 4.2\%. 
When our solver is used only during inference, we still see a 3.0\% increase in RR. 
Even though the performance of our \textit{Q-REG} decreases by 1.6\% in RR without LO, it provides a good initial pose and the performance gap can be easily bridged with a refinement step. For RANSAC 50K, the increase in RR is only 0.4\% after applying local optimization, indicating that many of the initially predicted poses is unreasonable and cannot be improved with further refinement. We can also observe a noticeable difference in performance between GeoTr + RL and GeoTr + QL, which further highlights the superiority of our quadric fitting approach.
When considering the mean RRE, RTE, and RMSE, we observe that our self-baselines provide consistently more robust results over all valid pairs versus the standard GeoTr (standard GeoTr corresponds to the top two rows in Table~\ref{tab:abl_3dlomatch}).
The ablation on the \textit{3DMatch} dataset is in the supplementary.

\textbf{Run-time.} We compute the average run-time in seconds per component in Table~\ref{tab:abl_runtime} (evaluated with GeoTr on \textit{3DLoMatch}). Regarding RANSAC 50K, which yields at least 2\% lower RR, \textit{QREG} provides better results while being an order of magnitude faster. 
On average, GeoTr's correspondence matcher  runs for $0.134$s. 
The overall inference time of each method can be obtained by adding it to Table~\ref{tab:abl_runtime}. 
These experiments were run on 8 Intel Xeon Gold 6150 CPUs and an NVIDIA GeForce RTX 3090 GPU.

\begin{table}[t]
    \footnotesize
    \begin{center}
    \resizebox{\columnwidth}{!}{
    \begin{tabular}{l|ccc|ccc}
        \toprule
        \multirow{2}{*}{Model} & RR & RRE & RTE & \multicolumn{3}{c}{\textit{Mean}} \\
         & ($\%$)$\uparrow$ & ($^{\circ}$)$\downarrow$ & (cm)$\downarrow$ & RRE $\downarrow$ & RTE $\downarrow$ & RMSE $\downarrow$ \\
	    \midrule
	    GeoTr + LGR  & 74.1 & 2.59 & \underline{7.3} & 23.15 & 58.3 & 57.8\\
	    GeoTr + 50K  & 75.0 & 2.54 & 7.7 & 22.69 & 57.8 & 57.3 \\
        \midrule
	    i) GeoTr + WK~\cite{umeyama1991least}         & 58.6 & 3.01 & 8.8 & 33.74 & 84.7 & 76.1\\
	    ii) GeoTr + ICP~\cite{besl1992method} & 75.1 & \underline{2.43} & 8.1 & 22.68 & 66.5 & 66.5 \\
        iii) GeoTr + PointDSC~\cite{bai2021pointdsc} & 74.0 & 2.55 & 7.5 & 23.95 & 61.6 & 60.7 \\
        iv) GeoTr + SC2-PCR~\cite{chen2022sc2} & 74.2 & 2.58 & 7.5 & 22.90 & 59.1 & 58.4 \\
        v) GeoTr + Q-REG w/ PCA  & 75.1 & 2.44 & 7.6 & 22.66 & 57.4 & 56.9 \\
	    vi) GeoTr + Q-REG w/ PCD         & 76.5 & 2.47 & 7.5 & 16.81 & 46.4 & \underline{44.6} \\
	    vii) GeoTr + Q-REG  & \underline{77.1} & 2.44 & 7.7 & \underline{16.70} & \textbf{46.0} & \underline{44.6} \\
        viii) GeoTr + \textit{\textbf{Q-REG*}} & \textbf{78.3} & \textbf{2.38} & \textbf{7.2} & \textbf{15.65} & \underline{46.3} & \textbf{42.5} \\
    	\bottomrule
    \end{tabular}
    }
    \end{center}
    \vspace{-5mm}
\caption{Results on the \textbf{\textit{3DLoMatch}}~\cite{predator} dataset of GeoTr~\cite{qin2022geometric} with different estimators. The best values are \textbf{bold} and the 2nd best are \underline{underlined}.}
\label{tab:tbd_3dlomatch}
\end{table}

\begin{table}[t]
    \footnotesize
    \begin{center}
    \resizebox{\columnwidth}{!}{
    \begin{tabular}{l|ccc|ccc}
        \toprule
        \multirow{2}{*}{Model} & RR & RRE & RTE & \multicolumn{3}{c}{\textit{Mean}} \\
         & ($\%$)$\uparrow$ & ($^{\circ}$)$\downarrow$ & (cm)$\downarrow$ & RRE $\downarrow$ & RTE $\downarrow$ & RMSE $\downarrow$ \\
	    \midrule
	    GeoTr + LGR  & 74.1 & 2.59 & \underline{7.3} & 23.15 & 58.3 & 57.8\\
	    GeoTr + 50K  & 75.0 & 2.54 & 7.7 & 22.69 & 57.8 & 57.3 \\
        \midrule
	    i) GeoTr + Q          & 75.5 & 2.47 & 7.6 & 22.38 & 57.6 & 57.3\\
	    ii) GeoTr + QL (Q-REG) & 77.1 & 2.44 & 7.7 & \underline{16.70} & \textbf{46.0} & \underline{44.6} \\
            iii) GeoTr + RL  & 75.4 & 2.46 & 7.6 & 22.86 & 58.5 & 68.0 \\
	    iv) GeoTr + QT         & \underline{77.2} & \textbf{2.37} & 7.5 & 17.32 & 50.3 & 47.4 \\
	    v) GeoTr + QTL (\textit{\textbf{Q-REG*}}) & \textbf{78.3} & \underline{2.38} & \textbf{7.2} & \textbf{15.65} & \underline{46.3} & \textbf{42.5} \\
    	\bottomrule
    \end{tabular}
    }
    \end{center}
    \vspace{-5mm}
\caption{Ablation results on the \textbf{\textit{3DLoMatch}}~\cite{predator} dataset of  GeoTr~\cite{qin2022geometric} with different aspects of the \textit{Q-REG} solver. The best values are \textbf{bold} and the 2nd best are \underline{underlined}.}
\label{tab:abl_3dlomatch}
\end{table}

\begin{table}[t]
    \footnotesize
    \begin{center}
    \resizebox{0.8\columnwidth}{!}{
    \begin{tabular}{ccc|ccc}
        \toprule
        LGR & +1K & +50K & +Q & +QL (Q-REG) \\
	    \midrule
        0.016 & 0.053 & 1.809 & 0.085 & 0.166 \\
    	\bottomrule
    \end{tabular}
    }
    \end{center}
    \vspace{-5mm}
\caption[Run-time evaluation during inference]{Run-time evaluation in seconds during inference using GeoTr~\cite{qin2022geometric} on the \textbf{\textit{3DLoMatch}} dataset. 
Times shown for LGR, RANSAC running 1K and 50K iterations, Quadric solvers (Sec.~\ref{sec:solver}), and with the entire \textit{Q-REG} algorithm.  
}
\label{tab:abl_runtime}
\end{table}

%% file: sec/5_conclusion.tex
\section{Conclusion}
\label{sec:conclusion}

We present a novel solution for point cloud registration, \textit{Q-REG}, that utilizes rich geometric information to estimate the rigid pose from a single correspondence. 
It allows us to formalize the robust estimation as an exhaustive search and enable us to iterate through all the combinations and select the best rigid pose among them. 
It performs quick outlier rejection by filtering degenerate solutions and assumption inconsistent motions (\eg, related to scale). 
\textit{Q-REG} is agnostic to matching methods and is consistently improving their performance on all reported datasets, setting new state-of-the-art on these benchmarks.

%% file: sec/X_supplementary.tex
\clearpage
\setcounter{page}{1}
\maketitlesupplementary

\begin{abstract}
\noindent
In the supplemental material, we provide additional details about the following:
\begin{enumerate}[leftmargin=12pt,itemsep=0em]
\item A video that gives a summary of our method and results. This is in a separate file. 
\item Distribution plots of results for the registration metrics RRE, RTE, and RMSE (Section~\ref{sec:plots}).
\item Additional qualitative results for the 3DLoMatch, 3DMatch and KITTI datasets (Section~\ref{sec:qua_results}).
\item Additional experiments on 3DLoMatch, 3DMatch and ModelNet datasets (Section~\ref{sec:add_results}).
\item Evaluation metrics illustrations (Section~\ref{metrics}).
\end{enumerate}

\end{abstract} 

\section{Cumulative Distribution Functions}
\label{sec:plots}

\begin{figure*}[!t]
    \centering
    \includegraphics[width=0.99\linewidth]{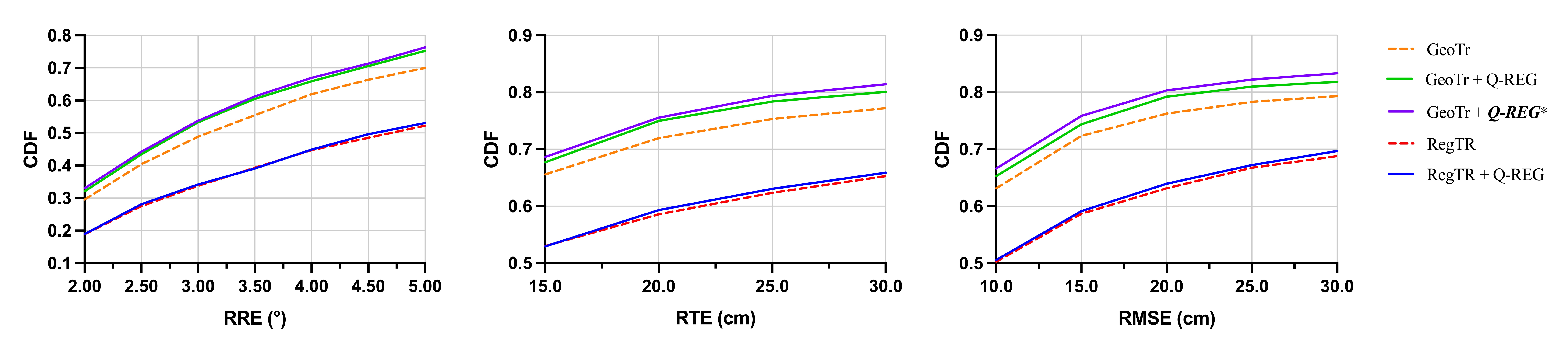}
    \caption{\textbf{Cumulative distribution functions of different registration metrics for the \textit{3DLoMatch~\cite{predator}} dataset.} When using Q-REG the performance of state-of-the-art matchers is increased, with the best results achieved when Q-REG is used for end-to-end training. GeoTr~\cite{qin2022geometric} refers to the standard formulation, GeoTr+Q-REG to using our method during inference, and GeoTr+\textbf{\textit{Q-REG*}} to using our method for end-to-end training. Similar for RegTr~\cite{yew2022regtr} and RegTr+Q-REG.
    Being close to the top-left corner is being more accurate.}
    \label{fig:3DLo_plots}
\end{figure*}

\begin{figure*}[!t]
    \centering
    \includegraphics[width=0.99\linewidth]{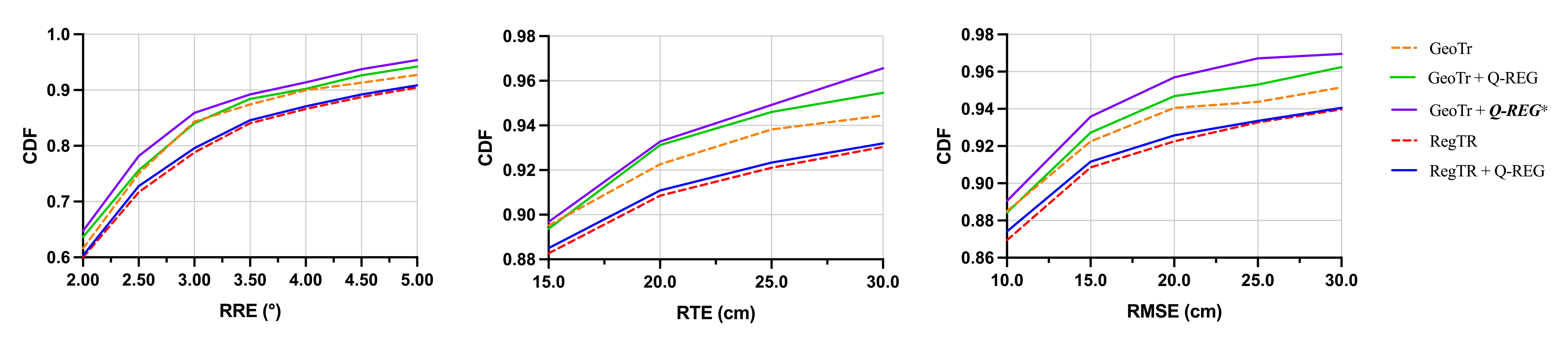}
    \caption{\textbf{Cumulative distribution functions of different registration metrics for the \textit{3DMatch~\cite{zeng20173dmatch}}.} When using Q-REG the performance of state-of-the-art matchers is increased, with the best results achieved when Q-REG is used for end-to-end training. GeoTr~\cite{qin2022geometric} refers to the standard formulation, GeoTr+\textit{Q-REG} to using our method during inference, and GeoTr+\textbf{\textit{Q-REG*}} to using our method for end-to-end training. Similar for RegTr~\cite{yew2022regtr} and RegTr+\textit{Q-REG}. 
    Being close to the top-left corner is interpreted as being accurate.}
    \label{fig:3D_plots}
\end{figure*}

\begin{figure*}[!t]
    \centering
    \includegraphics[width=0.75\linewidth]{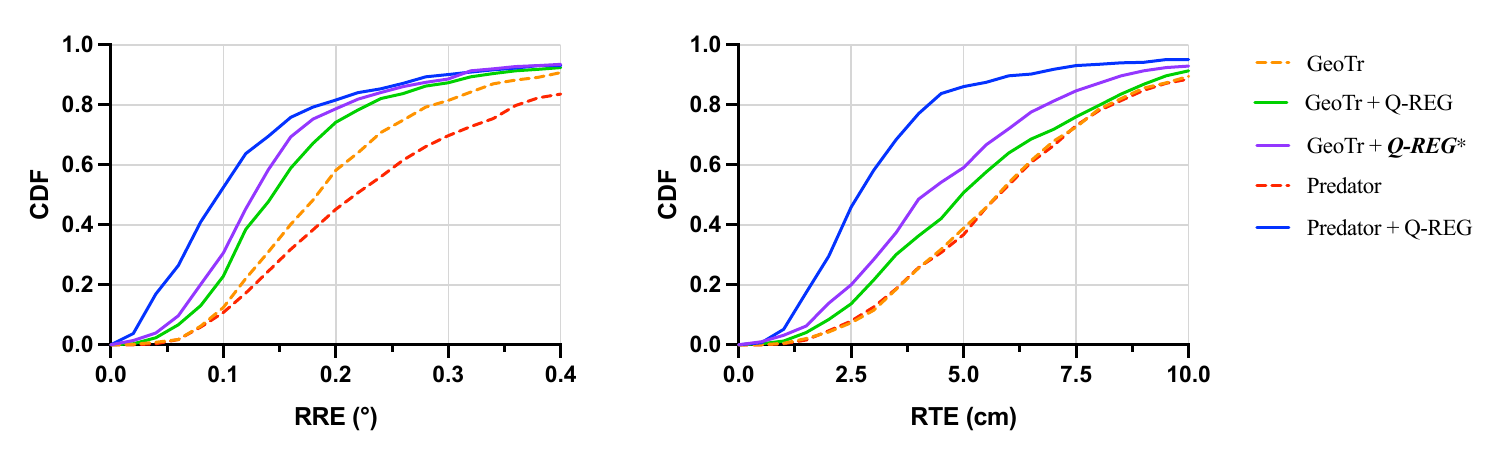}
    \caption{\textbf{Cumulative distribution functions of registration metrics on the \textit{KITTI~\cite{geiger2012we} dataset.}} When using Q-REG the performance of state-of-the-art matchers is increased, with the best results for GeoTr~\cite{qin2022geometric} achieved when Q-REG is used for end-to-end training. GeoTr refers to the standard formulation, GeoTr+\textit{Q-REG} to using our method during inference, and GeoTr+\textbf{\textit{Q-REG*}} to using our method for end-to-end training. Similar for Predator~\cite{predator} and Predator+\textit{Q-REG}. 
    Being close to the top-left corner is interpreted as being accurate.}
    \label{fig:KITTI_plots}
\end{figure*}

In this section, we plot the Cumulative Distribution Functions (CDFs) of the following registration metrics: Relative Rotation Error (RRE), Relative Translation Error (RTE), and Root Mean Square Error (RMSE), on the \textit{3DLoMatch}~\cite{predator} (Figure~\ref{fig:3DLo_plots}) and \textit{3DMatch}~\cite{zeng20173dmatch} (Figure~\ref{fig:3D_plots}) datasets. 
We also plot the CDFs of RRE and RTE on the \textit{KITTI}~\cite{geiger2012we} (Figure~\ref{fig:KITTI_plots}) dataset. 
Being close to the top-left corner is interpreted as being accurate. 
As expected, when using Q-REG, state-of-the-art correspondence matching algorithms have improved performance with respect to their standard formulation.
When using the matcher trained in an end-to-end fashion with \textbf{\textit{Q-REG*}}, the performance can be further improved. 
This stands for all datasets. 

\section{Qualitative results}
\label{sec:qua_results}

In Figures~\ref{fig:qual_3DLo} and \ref{fig:qual_3D}, we showcase additional qualitative registration results for the \textit{3DLoMatch} and \textit{3DMatch} datasets, respectively. We evaluate the top-2 performing state-of-the-art matchers on these datasets, namely GeoTr~\cite{qin2022geometric} and RegTr~\cite{yew2022regtr}. Please note that GeoTr Standard (d-i) refers to the standard formulation, GeoTr+\textit{Q-REG} (d-ii) to using our method during inference, and GeoTr+\textbf{\textit{Q-REG*}} (d-iii) to using our method for end-to-end training. 
The same holds for RegTr Standard (c-i) and RegTr+\textit{Q-REG} (c-ii). 
In Figure~\ref{fig:qual_KITTI}, we present qualitative registration results for the \textit{KITTI} dataset. We evaluate the state-of-the-art matchers, GeoTr and Predator~\cite{predator}, on the dataset. The notations for GeoTr are the same. Predator (c-i) refers to the standard formulation and Predator+\textit{Q-REG} (c-ii) refers to using our method during inference.

\subsection{3DLoMatch Dataset}
\label{sec:3dlomatch}

\begin{figure*}[p]
    \centering
    \includegraphics[width=0.99\linewidth]{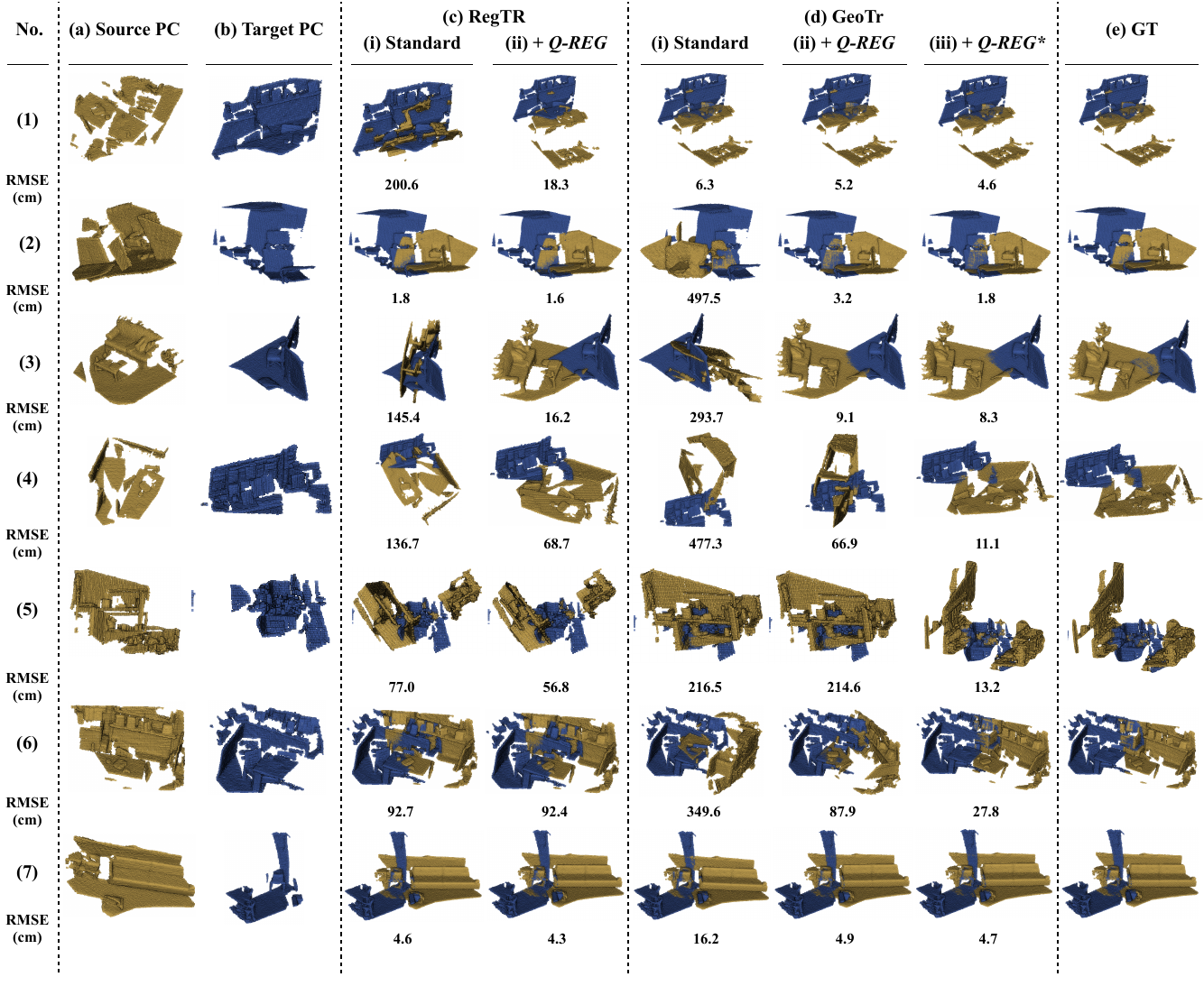}
    \caption{\textbf{Qualitative Results for the \textit{3DLoMatch~\cite{predator}} dataset.} See Section~\ref{sec:3dlomatch} for an explanation of the results. \textit{Best viewed in screen.}}
    \label{fig:qual_3DLo}
\end{figure*}

In Figure~\ref{fig:qual_3DLo}, we illustrate examples of point cloud registration for the \textit{3DLoMatch} dataset. We also report the RMSE for all results, which we use to determine if two point clouds are correctly registered. Specifically, per row:

\vspace{1mm}\noindent\textbf{Row (1):} In the case of GeoTr, the standard formulation (d-i) already produces well-aligned point clouds, and the addition of Q-REG (d-ii and d-iii) slightly refines this output. However, in the case of RegTr, we can see the most improvement. The standard formulation (c-i) fails to achieve a good alignment and Q-REG (c-ii) is able to recover a good pose. This means that our method is able to identify robust correspondences and remove spurious ones.

\vspace{1mm}\noindent\textbf{Row (2):} In this example, we observe the opposite behavior, where Reg-Tr+\textit{Q-REG} (c-ii) further refines the good results achieved by RegTr standard (c-i). GeoTr standard (d-i) fails to achieve a good registration, but the use of Q-REG (d-ii and d-iii) is able to recover the pose.

\vspace{1mm}\noindent\textbf{Row (3):} Here, both RegTr standard (c-i) and GeoTr standard (d-i) fail to align the point clouds correctly. Despite this failure, in both matchers, the use of Q-REG (c-ii, d-ii, and d-iii) allows to recover the final pose. 
    
\vspace{1mm}\noindent\textbf{Rows (4) and (5):} In these examples, the only method that allows to recover an accurate pose is GeoTr+\textbf{\textit{Q-REG*}} (d-iii). This means that using Q-REG in end-to-end training can provide additional improvements in performance by learning how to better match correspondences together with the objective of rigid pose estimation, and not in isolation as in all other cases (c-i, c-ii, d-i, \& d-ii).

\vspace{1mm}\noindent\textbf{Row (6):} In this example, all methods fail. However, in the case of GeoTr, we can see noticeable improvement using Q-REG. The pose can be further recovered with \textbf{\textit{Q-REG*}}.
    
\vspace{1mm}\noindent\textbf{Row (7):} In this final example, all methods perform reasonably well. However, in the case of GeoTr+\textit{Q-REG} (d-ii) and GeoTr+\textbf{\textit{Q-REG*}} (d-iii), we are able to reduce the RMSE from 16 cm in the standard formulation (d-i) to 5 cm, which is a substantial improvement in the estimated pose.

\subsection{3DMatch Dataset}
\label{sec:3dmatch}

\begin{figure*}[p]
    \centering
    \includegraphics[width=0.99\linewidth]{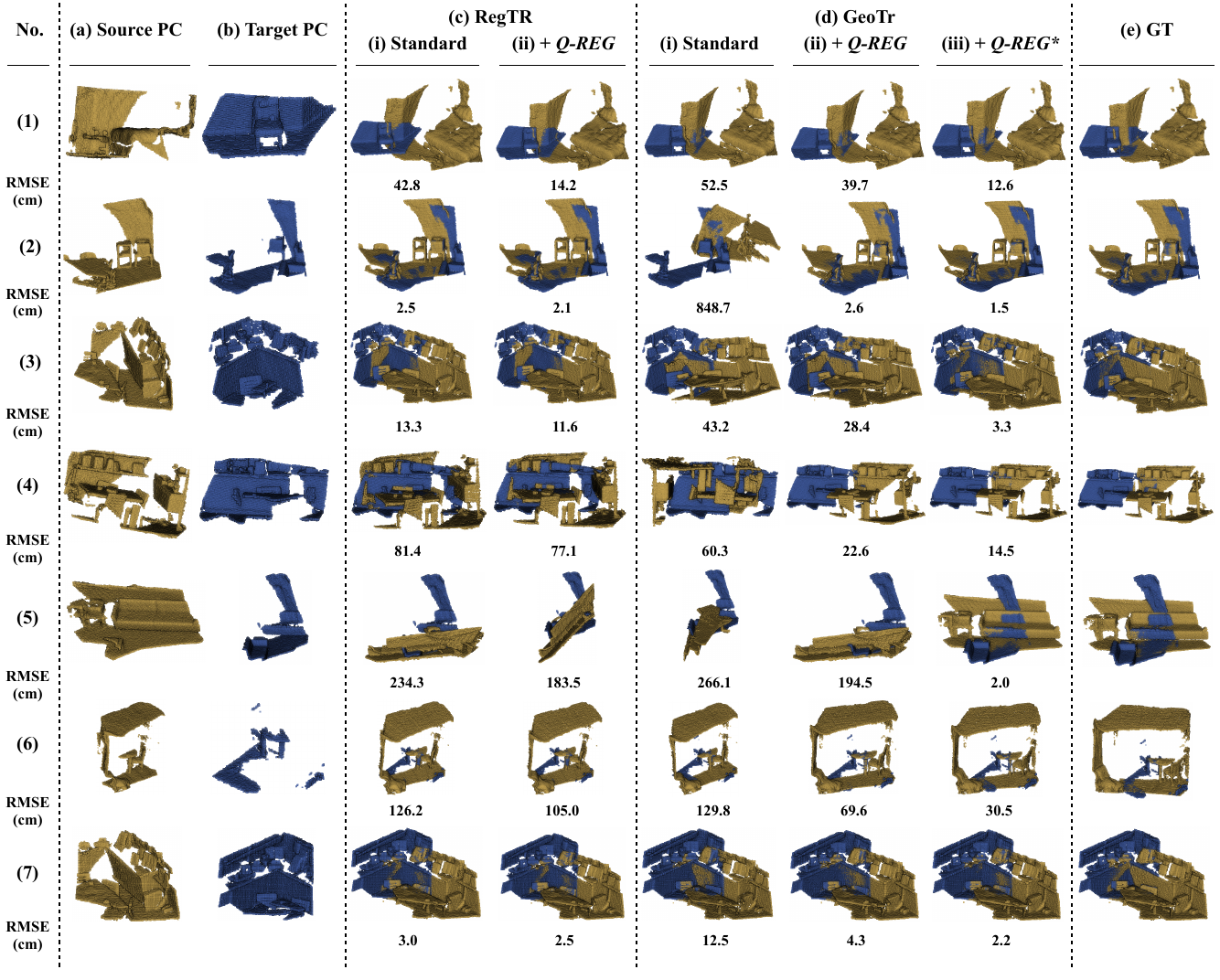}
    \caption{\textbf{Qualitative Results for the \textit{3DMatch~\cite{zeng20173dmatch}} dataset.} See Section~\ref{sec:3dmatch} for an explanation of the results. \textit{Best viewed in screen.}}
    \label{fig:qual_3D}
\end{figure*}

In Figure~\ref{fig:qual_3D}, we illustrate examples of point cloud registration for the \textit{3DMatch} dataset. Similarly, We report the RMSE for all results. Specifically, per row:

\vspace{1mm}\noindent\textbf{Rows (1):} In this example, we observe that the addition of Q-REG (c-ii) to the standard RegTr formulation (c-i) is able to greatly correct the estimated pose. In the case of GeoTr+\textit{Q-REG} (d-ii), we see that it cannot correct the error of GeoTr standard (c-i), however, GeoTr+\textbf{\textit{Q-REG*}} (d-iii) recovers the final pose. This points, as mentioned beforehand, to the power of learning to choose 
correspondences with the inclusion of the pose error minimization objective.

\vspace{1mm}\noindent\textbf{Rows (2):} Here, although we do not see a big improvement between RegTr standard (c-i) and RegTr+\textit{Q-REG} (c-ii), we can see it in the case of GeoTr. The standard formulation (d-i) fails to estimate a good alignment, but the use of Q-REG (d-ii) and \textbf{\textit{Q-REG*}} (d-iii) provide a close-to-GT pose estimation, with \textbf{\textit{Q-REG*}} being better than Q-REG.
    
\vspace{1mm}\noindent\textbf{Rows (3), (4) and (5):} In these examples, GeoTr standard (d-i) fails to achieve a good registration, but the use of Q-REG (d-ii) improves the rigid pose. And GeoTr+\textbf{\textit{Q-REG*}} further improves it and successfully recovers the final pose.

\vspace{1mm}\noindent\textbf{Rows (6):} In this example, all methods fail to recover a good pose. However, we can see a noticeable improvement with Q-REG in the case of GeoTr. The pose can be further recovered with \textbf{\textit{Q-REG*}}. 
    
\vspace{1mm}\noindent\textbf{Row (7):} In this final example, all methods perform reasonably well. However, in the case of GeoTr+\textit{Q-REG} (d-ii) and GeoTr+\textbf{\textit{Q-REG*}} (d-iii), we are able to substantially reduce the RMSE from 13 cm in the standard formulation (d-i) to 4 cm and 2 cm respectively.

\subsection{KITTI Dataset}
\label{sec:KITTI}
\begin{figure*}[p]
    \centering
    \includegraphics[width=0.95\linewidth]{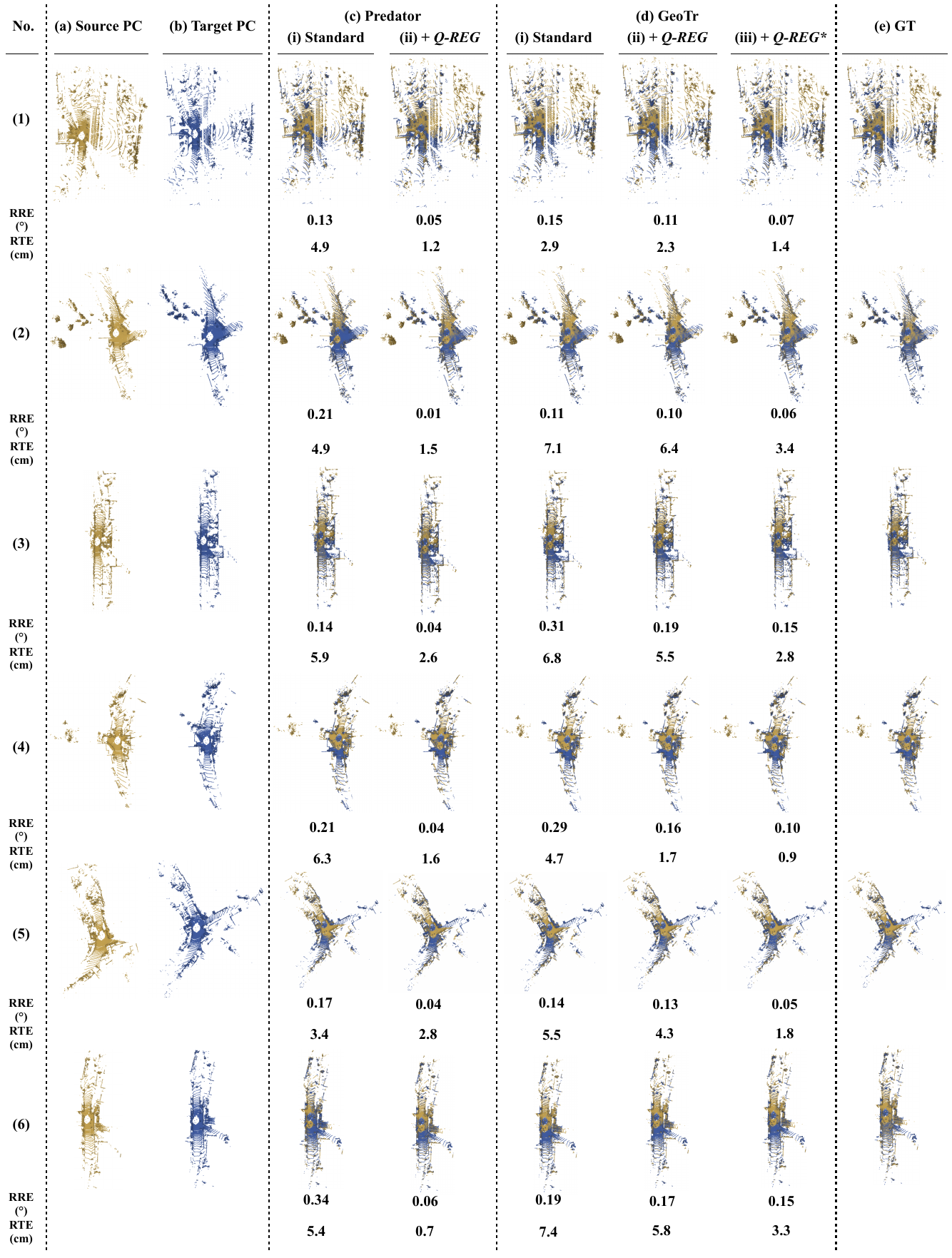}
    \caption{\textbf{Qualitative Results for the \textit{KITTI~\cite{geiger2012we}} dataset.} See Section~\ref{sec:KITTI} for an explanation of the results. \textit{Best viewed in screen.}}
    \label{fig:qual_KITTI}
\end{figure*}

In Figure~\ref{fig:qual_KITTI}, we show examples of point cloud registration for the \textit{KITTI} dataset. We also present RRE and RTE for all the registration results, which are the two criteria to compute the registration recall. As we can see, when using Q-REG, both GeoTr and Predator have improved performance with respect to their standard formulation. When GeoTr is trained in an end-to-end fashion with \textbf{\textit{Q-REG*}}, the performance can be further improved compared to Q-REG. 

\section{Additional Experiments}
\label{sec:add_results}

\subsection{Comparison with Other Estimators - 3DMatch}
\label{sec:comp_est_3D}

We compare \textit{Q-REG} with other estimators that predict rigid pose from correspondences on the \textit{3DMatch} dataset, using the state-of-the-art matcher GeoTr~\cite{qin2022geometric} as the correspondence extractor (best performing on this dataset). We evaluate the following estimators:\\ (i) \textbf{GeoTr + WK}: weighted variant of the Kabsch-Umeyama algorithm~\cite{umeyama1991least}. \\(ii) \textbf{GeoTr + ICP}: Iterative closest point (ICP)~\cite{besl1992method} initialized with 50K RANSAC. \\
(iii) \textbf{GeoTr + PointDSC}: PointDSC~\cite{bai2021pointdsc} with their pre-trained model from ~\cite{PointDSC_Github}.\\
(iv) \textbf{GeoTr + SC2-PCR}: GeoTr with SC2-PCR~\cite{chen2022sc2}.\\
(v) \textbf{GeoTr + Q-REG w/ PCA}: PCA instead of our quadric fitting to determine the local coordinate system. \\
(vi) \textbf{GeoTr + Q-REG w/ PCD}: Use principal direction as explained in Sec. 3.1. 
of the main paper. \\
(vii) \textbf{GeoTr + Q-REG}: Our quadric-fitting solver is used only in inference. \\
(viii) \textbf{GeoTr + \textit{Q-REG*}}: Our solver is used in both end-to-end training and inference. \\

The results are in Table~\ref{tab:tbd_3dmatch} (\textit{best results in \textbf{bold}, second best are \underline{underlined}}). Among all methods, \textbf{\textit{Q-REG*}} performs the best in the majority of the metrics. \\
(i) GeoTr + WK shows a large performance gap with other methods since it utilizes soft correspondences and is not robust enough to outliers. \\
(ii) GeoTr + ICP relies heavily on the initialization and, thus, often fails to converge to a good solution. \\
(iii) GeoTr + PointDSC has comparable results as LGR but is noticeably worse than \textit{Q-REG}. \\
(iv) GeoTr + Q-REG w/ PCA leads to less accurate results than w/ quadric fitting, which demonstrates the superiority of \textit{Q-REG} that utilizes quadrics. \\
(v) GeoTr + Q-REG w/ PCD shows comparable performance as our \textit{Q-REG}. This is expected since both methods have the same geometric meaning. However, there is a substantial difference in runtime (secs): 0.166 for quadric fitting versus 0.246 for PCD.

 \begin{table}[h]
    \footnotesize
    \begin{center}
    \resizebox{\columnwidth}{!}{
    \begin{tabular}{l|ccc|ccc}
        \toprule
        \multirow{2}{*}{Model} & RR & RRE & RTE & \multicolumn{3}{c}{\textit{Mean}} \\
         & ($\%$)$\uparrow$ & ($^{\circ}$)$\downarrow$ & (cm)$\downarrow$ & RRE $\downarrow$ & RTE $\downarrow$ & RMSE $\downarrow$ \\
	    \midrule
	   GeoTr + LGR  & 92.5 & 1.54 & \textbf{5.1} & 7.04 & 19.4 & 17.6 \\
	    GeoTr + 50K  & 92.2 & 1.66 & 5.6 & 6.85 & 18.7 & 17.1\\
        \midrule
	    i) GeoTr + WK~\cite{umeyama1991least}          & 86.5 & 1.69 & 5.5 & 9.34 & 27.8 & 22.8 \\
	    ii) GeoTr + ICP~\cite{besl1992method} & 91.7 & \textbf{1.52} & 5.3 & 5.76 & 18.5 & 16.8 \\
        iii) GeoTr + PointDSC~\cite{bai2021pointdsc} & 91.8 & 1.55 & \textbf{5.1} & 7.10 & 20.6 & 18.7 \\
        iv) GeoTr + SC2-PCR~\cite{chen2022sc2} & 92.3 & 1.56 & 5.2 & 7.18 & 20.5 & 18.6 \\
        v) GeoTr + Q-REG w/ PCA  & 91.9 & 1.55 & 5.3 & 6.65 & 18.0 & 16.5 \\
	    vi) GeoTr + Q-REG w/ PCD         & 93.6 & 1.55 & \underline{5.2} & \underline{4.73} & \underline{14.7} & \underline{12.4} \\
	    vii) GeoTr + Q-REG  & \underline{93.8} & 1.57 & 5.3 & 4.74 & 15.0 & 12.8 \\
        viii) GeoTr + \textit{\textbf{Q-REG*}} & \textbf{95.2} & \underline{1.53} & 5.3 & \textbf{3.70} & \textbf{12.5} & \textbf{10.7} \\
    	\bottomrule
    \end{tabular}
    }
    \end{center}
\caption{Results on the \textbf{\textit{3DMatch}}~\cite{predator} dataset of GeoTr~\cite{qin2022geometric} with different estimators. The best values are \textbf{bold} and the 2nd best are \underline{underlined}.}
\label{tab:tbd_3dmatch}
\end{table}

\subsection{Ablation Study of Q-REG on 3DMatch}
\label{sec:ablation_3D}

We perform ablation studies to evaluate the contribution of each component in the \textit{Q-REG} solver to the best-performing matcher on the \textit{3DMatch} datasets, the state-of-the-art GeoTr~\cite{qin2022geometric}. We evaluate the following self-baselines (i, ii and iii are inference only): \\
(i) \textbf{GeoTr + Q}: Our quadric-fitting 1-point solver. \\
(ii) \textbf{GeoTr + QL}: We extend the quadric fitting with local optimization (LO) as discussed in Sec. 3.3. End-to-End Training of the main paper.\\
(iii) \textbf{GeoTr + RL}: We replace quadric fitting with RANSAC 50K in (ii).\\
(iv) \textbf{GeoTr + QT}: Our quadric-fitting solver is used in end-to-end training, 
while during inference we do not employ LO.\\
(v) \textbf{GeoTr + QTL}: Our quadric-fitting 1-point solver is used in end-to-end training followed by inference using LO. \\

The results are reported in Table~\ref{tab:abl_3dmatch} (\textit{best results in \textbf{bold}, second best are \underline{underlined}}). 
\textbf{\textit{Q-REG*}} performs the best in the majority of the metrics. 
Specifically, we observe that there is a substantial increase in RR by 3.0\%. 
When our solver is used only during inference, we still see a 1.6\% increase in RR. 
Even though the performance of our \textit{Q-REG} decreases by 1.2\% in RR without LO, it provides a good initial pose and the performance gap can be easily bridged with a refinement step. For RANSAC 50K, the increase in RR is only 0.2\% after applying local optimization, indicating that many of the initially predicted poses is unreasonable and cannot be improved with further refinement. We can also observe a noticeable difference in performance between GeoTr + RL and GeoTr + QL, which further highlights the superiority of our quadric fitting approach.
When considering the mean RRE, RTE, and RMSE, we observe that our self-baselines provide consistently more robust results over all valid pairs versus the standard GeoTr (standard GeoTr corresponds to the top two rows in Table~\ref{tab:abl_3dmatch}).

   \begin{table}[h]
    \footnotesize
    \begin{center}
    \resizebox{\columnwidth}{!}{
    \begin{tabular}{l|ccc|ccc}
        \toprule
        \multirow{2}{*}{Model} & RR & RRE & RTE & \multicolumn{3}{c}{\textit{Mean}} \\
         & ($\%$)$\uparrow$ & ($^{\circ}$)$\downarrow$ & (cm)$\downarrow$ & RRE $\downarrow$ & RTE $\downarrow$ & RMSE $\downarrow$ \\
	    \midrule
	    GeoTr + LGR  & 92.5 & 1.54 & \textbf{5.1} & 7.04 & 19.4 & 17.6 \\
	    GeoTr + 50K  & 92.2 & 1.66 & 5.6 & 6.85 & 18.7 & 17.1\\  
        \midrule
	    i) GeoTr + Q                             & 92.6 & 1.55 & 5.3 & 6.26 & 17.4 & 15.8 \\
	    ii) GeoTr + QL (Q-REG)            & 93.8 & 1.57 & 5.3 & 4.74 & 15.0 & 12.8 \\
            iii) GeoTr + RL  & 92.4 & 1.55 & \underline{5.2} & 6.81 & 18.6 & 17.4 \\
	    iv) GeoTr + QT                            & \underline{94.3} & \textbf{1.51} & \underline{5.2} & \underline{3.78} & \underline{12.8} & \underline{10.9} \\
	    v) GeoTr + QTL (\textit{\textbf{Q-REG*}}) & \textbf{95.2} &  \underline{1.53} & 5.3 & \textbf{3.70} & \textbf{12.5} & \textbf{10.7} \\
	    \bottomrule
    \end{tabular}
    }
    \end{center}
    \caption{Ablation results on the \textbf{\textit{3DMatch}}~\cite{zeng20173dmatch} dataset of GeoTr~\cite{qin2022geometric} with different aspects of the \textit{Q-REG} solver. The best values are \textbf{bold} and the 2nd best are \underline{underlined}.}
    \label{tab:abl_3dmatch}
\end{table}

\subsection{Comparison on ModelNet}

Qin et al.~\cite{qin2022geometric} do not provide any \textit{ModelNet} dataset evaluation in their paper, however, they offer an evaluation on their GitHub page \cite{GeoTr_Github}. 
We follow their proposed protocol and dataset split for the standard setting. In a nutshell, (i) they use a portion of the \textit{ModelNet} dataset that excludes the axis-symmetrical object categories and (ii) they train and test on all categories, instead of keeping certain ones unseen for testing. We trained GeoTr~\cite{qin2022geometric} from scratch on this setting and report the results on the same metrics in Table~\ref{tab:modelnet45} (\textit{best results marked in \textbf{bold}}). We note that our solver, in both an only-inference setting and an end-to-end-training one, performs the best, and minimizes all three metrics with respect to the best-performing GeoTr without \textit{Q-REG}.

\begin{table}[h]
    \footnotesize
    \begin{center}
    \resizebox{\columnwidth}{!}{
    \begin{tabular}{l|ccc}
        \toprule
        Method & CD $\downarrow$ & RRE ($^{\circ}$)$\downarrow$ & RTE (cm)$\downarrow$ \\
	    \midrule
        GeoTr~\cite{qin2022geometric}                            & 0.00093 & 1.61 & 1.9\\
	    GeoTr~\cite{qin2022geometric} + 50K                      & 0.00112 & 1.94 & 2.3\\
	    GeoTr~\cite{qin2022geometric} + Q-REG           & {0.00088} & {1.43} & {1.7}\\
	    GeoTr~\cite{qin2022geometric} + \textbf{\textit{Q-REG*}} & \textbf{0.00085} & \textbf{1.26} & \textbf{1.5}\\
    	\bottomrule
    \end{tabular}
    }
    \end{center}
\caption{Evaluation of GeoTr~\cite{qin2022geometric} on a different \textbf{\textit{ModelNet}} setting: (i) using a portion of it that excludes axis-symmetrical categories and (ii) using all categories in both training and testing.}
\label{tab:modelnet45}
\end{table}

\subsection{Inlier Ratio (IR) and Feature Matching Recall (FMR)}

\begin{figure*}[p]
    \centering
    \includegraphics[width=0.95\linewidth]{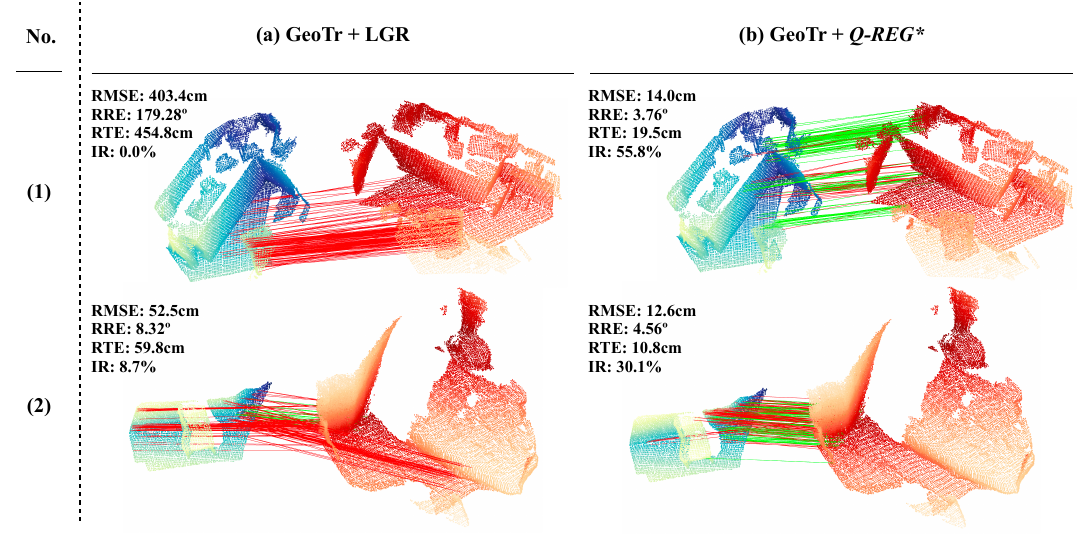}
    \caption{Matching on the \textit{\textbf{3DMatch}} and \textit{\textbf{3DLoMatch}} datasets.}
    \label{fig:matching}
\end{figure*}

We report the IR and FMR of GeoTr w/wo \textbf{\textit{Q-REG*}} on the \textit{3DMatch} and \textit{3DLoMatch} datasets in Table~\ref{tab:metrics}. 
As we can see, training GeoTr end-to-end using the proposed \textbf{\textit{Q-REG*}} method leads to substantial improvements both in terms of IR and FMR. This means that our \textbf{\textit{Q-REG*}} method can not only boost the registration results but also improve the matching results. We also show visual matching results in Figure~\ref{fig:matching} of the original GeoTr and when it is end-to-end trained with \textbf{\textit{Q-REG*}}. 
GeoTr+\textbf{\textit{Q-REG*}} substantially improves the matching and registration results.

\begin{table}[t]
    \footnotesize    
    \begin{center}
    \resizebox{\columnwidth}{!}{
    \begin{tabular}{l|l|cc}
        \hline
        \multirow{1}{*}{\textbf{Dataset}} & \multirow{1}{*}{\textbf{Model}} & \textbf{IR} (\%$\uparrow$) & \textbf{FMR} (\%$\uparrow$) \\
	    \hline
	    \multirow{2}{*}{3DMatch} & GeoTr   & 70.9 & 98.2 \\ 
             & GeoTr + \textbf{\textit{Q-REG*}}   & \textbf{78.1} & \textbf{98.7} \\ 
            \hline
	    \multirow{2}{*}{3DLoMatch} & GeoTr   & 43.5 & 87.1 \\ 
             & GeoTr + \textbf{\textit{Q-REG*}}   & \textbf{50.1} & \textbf{88.2} \\ 
    	\hline
    \end{tabular}
    }
    \end{center}
\caption{  
IR and FMR of GeoTr w/wo \textbf{\textit{Q-REG*}} on the \textbf{\textit{3DMatch}} and \textbf{\textit{3DLoMatch}} datasets. Best values per group in \textbf{bold}. 
}
\label{tab:metrics}
\end{table}

\section{Evaluation Metrics}
\label{metrics}

We follow the standard evaluation metrics per dataset and we provide here their description.

\subsection{3DMatch \& 3DLoMatch}
\textit{Relative Rotation Error} (RRE) is the geodesic distance in degrees between the rotation matrices that represent the estimated rotation and the ground-truth rotation.
\begin{equation}
    \label{eq:RRE}
    \text{RRE} = \mathrm{arccos} \left( \frac{\mathrm{trace}(\mathbf{R}^T \cdot \bar{\mathbf{R}}-1)}{2} \right).
\end{equation}

\textit{Relative Translation Error} (RTE) is the Euclidean distance between estimated and ground-truth translation. 
\begin{equation}
    \label{eq:RTE}
    \text{RTE} = \Vert \mathbf{t}-\bar{\mathbf{t}} \Vert_2^2.
\end{equation}

\textit{Registration Recall} (RR) is defined as the fraction of the point cloud pairs whose transformation error is smaller than 0.2m, which is the most reliable metric. The transformation error is defined as the root mean square error of the ground-truth correspondences $\mathcal{C}^*$ after applying the estimated transformation $\mathbf{T}_{\mathbf{P} \to \mathbf{Q}}$:
\begin{equation}
    \label{eq:RMSE}
    \text{RMSE} = \sqrt{\frac{1}{\left|\mathcal{C}^*\right|}\sum_{\left({\Point}^*_{x_i}, {\mathbf{q}}^*_{y_i}\right) \in \mathcal{C}^*} \Vert \mathbf T_{\mathbf{P} \to \mathbf{Q}} \left(\mathbf{\Point}^*_{x_i}\right)- \mathbf{q}^*_{y_i} \Vert_2^2 },
\end{equation}
\begin{equation}
    \label{eq:RR}
     \text{RR}=\frac{1}{M}\sum_{i=1}^{M} \llbracket \text{RMSE}_i<0.2m \rrbracket.
\end{equation}

The mean RRE, RTE, and RMSE in the main paper are evaluated over all valid pairs\footnote{According to~\cite{zeng20173dmatch}, a valid pair is a pair of non-consecutive frames.} instead of only those with an RMSE below 0.2 m, and we provide a simple average over all valid pairs instead of the median value of each scene followed by the average over all scenes. These metrics will show how consistently well (or not) a method performs in registering scenes.

\subsection{KITTI}

\textit{Relative Rotation Error} (RRE) and \textit{Relative Translation Error} (RTE) are similarly defined on this dataset.

\textit{Registration Recall} (RR) is the fraction of the point cloud pairs whose RRE and RTE are both below certain thresholds (\ie, RRE $<5^{\circ}$ and RTE $<2$m).
\begin{equation}
    \label{eq:KITTI_RR}
    \text{RR} = \frac{1}{\mathnormal{M}}\sum_{i=1}^\mathnormal{M} \llbracket \text{RRE}_i < 5^{\circ} \land \text{RTE}_i < 2m \rrbracket.
\end{equation}

Following~\cite{bai2020d3feat, Choy2019FCGF, predator, lu2021hierarchical, yu2021cofinet, qin2022geometric}, the mean RRE and RTE are only evaluated over the correctly registered point cloud pairs.

\subsection{ModelNet \& ModelLoNet}
\label{cham_dis}

\noindent
\textit{Relative Rotation Error} (RRE) and \textit{Relative Translation Error} (RTE) are similarly defined on this dataset.

Following~\cite{yew2020rpm, predator}, we use the \textit{modified} chamfer distance (CD) metric:
\begin{equation}
\label{eq:predator_CD}
\begin{split}
    \Tilde{CD}\left(\mathbf{P},\mathbf{Q} \right) =& \frac{1} {\left|\mathbf{P}\right|} \sum_{\Point \in \mathbf{P}} \min_{\mathbf{q} \in \mathbf{Q}_\text{raw}} \Vert \mathbf{T}^\mathbf{Q}_\mathbf{P}\left(\Point\right) - \mathbf{q}\Vert_2^2 + \\ & \frac{1} {\left|\mathbf{Q}\right|} \sum_{\mathbf{q} \in \mathbf{Q}} \min_{\Point \in \mathbf{P}_\text{raw}} \Vert \mathbf{q} - \mathbf{T}^\mathbf{Q}_\mathbf{P}\left(\Point\right) \Vert_2^2
\end{split}
\end{equation}
where $\mathbf{P}_\text{raw}\in\mathbb{R}^{2048 \times3}$ and $\mathbf{Q}_\text{raw}\in\mathbb{R}^{2048 \times3}$ are \textit{raw} source and target point clouds, $\mathbf{P}\in\mathbb{R}^{717 \times3}$ and $\mathbf{Q}\in\mathbb{R}^{717 \times3}$ are \textit{input} source and target point clouds.